\documentclass[journal]{IEEEtran}
\usepackage{cite}

\ifCLASSINFOpdf
   \usepackage[pdftex]{graphicx}
   \usepackage{subfigure}
   \graphicspath{{../pdf/}{../jpeg/}}
   \DeclareGraphicsExtensions{.pdf,.jpeg,.png}
\else
   \usepackage[dvips]{graphicx}
   \usepackage{subfigure}
   \graphicspath{{../eps/}}
   \DeclareGraphicsExtensions{.eps}
\fi

\usepackage[ruled,linesnumbered]{algorithm2e}
\usepackage{color}
\makeatletter

\newcommand{\Rmnum}[1]{\expandafter\@slowromancap\romannumeral #1@}
\makeatother

\usepackage{amsmath}
\usepackage{amsfonts}       
\usepackage{bm}
 \usepackage{algorithmic}
\usepackage{array}

\ifCLASSOPTIONcompsoc
  \usepackage[caption=false,font=normalsize,labelfont=sf,textfont=sf]{subfig}
\else
  \usepackage[caption=false,font=footnotesize]{subfig}
\fi

\usepackage{stfloats}

\usepackage{multirow}

\begin{document}
%
\title{A Bioinspired Approach-Sensitive Neural Network for Collision Detection in Cluttered and Dynamic Backgrounds}
%
%
%

\author{Xiao~Huang, Hong~Qiao,~\IEEEmembership{Fellow,~IEEE}, Hui~Li, and~Zhihong~Jiang
\thanks{X. Huang, H. Li and Z. H. Jiang are with the School of Mechatronical Engineering, Advanced Innovation Center for Intelligent Robots and Systems, Key Laboratory of Biomimetic Robots and Systems of Chinese Ministry of Education, Beijing Institute of Technology, Beijing, 100081, China (e-mail:7520200120@bit.edu.cn; lihui2011@bit.edu.cn; jiangzhihong@bit.edu.cn).}
\thanks{H. Qiao is with the State Key Laboratory of Management and Control for Complex Systems, Institute of Automation, Chinese Academy of Sciences, Beijing, 100190, China, is also with Beijing Key Laboratory of Research and Application for Robotic Intelligence of 'Hand-Eye-Brain' Interaction, Beijing, 100190 and the University of Chinese Academy of Sciences, Beijing, 100049, e-mail: (hong.qiao@ia.ac.cn).}
\thanks{This work is supported in part by the National Key Research and Development Program of China under 2018YFB1305300, 2017YFB1300200 and 2017YFB1300203, the China Postdoctoral Science Foundation under Grant 2020TQ0039, the National Natural Science Foundation of China under Grant 61733001, U2013602, 61873039, U1913211 and U1713215. (Corresponding author: Hui Li, Zhihong Jiang)}
}

\maketitle


\begin{abstract}
Rapid, accurate and robust detection of looming objects in cluttered moving backgrounds is a significant and challenging problem for robotic visual systems to perform collision detection and avoidance tasks. Inspired by the neural circuit of elementary motion vision in the mammalian retina, this paper proposes a bioinspired approach-sensitive neural network (ASNN) that contains three main contributions. Firstly, a direction-selective visual processing module is built based on the spatiotemporal energy framework, which can estimate motion direction accurately via only two mutually perpendicular spatiotemporal filtering channels. Secondly, a novel approach-sensitive neural network is modeled as a push-pull structure formed by ON and OFF pathways, which responds strongly to approaching motion while insensitivity to lateral motion. Finally, a method of directionally selective inhibition is introduced, which is able to suppress the translational backgrounds effectively. Extensive synthetic and real robotic experiments show that the proposed model is able to not only detect collision accurately and robustly in cluttered and dynamic backgrounds but also extract more collision information like position and direction, for guiding rapid decision making.

\end{abstract}


%
\IEEEpeerreviewmaketitle

\section{INTRODUCTION}
%
%
%
%

\IEEEPARstart{R}{apid} and robust detection of approaching objects, such as looming predators, is essential for animals to survive in nature. It is also of critical importance and challenging for robotic systems, such as ground vehicles, UAVs and other robots, to perform collision detection and obstacle avoidance. Currently, collision-detecting systems mainly rely on infrared sensors, radar, laser, ultrasound, optical sensors, and multi-sensor fusion. Among them, vision-based detection is becoming popular due to rich input information, low cost, biological plausibility, and good flexibility in different environments. Plenty of visual motion perception strategies have been proposed to improve the performance of collision detection. However, it is still an open challenge for robotic visual systems to detect collisions accurately and robustly and to evoke reflex decisions rapidly, especially in cluttered and highly dynamic environments. This is because, in the absence of distance information, the moving background is bound to cause huge interference to the recognition of the looming targets. Meanwhile, more motion information, like the position and velocity of the colliding target, is usually required timely to guide subsequent decisions.

Over the past decades, many conventional computer vision methods have emerged for real-time collision detection. As summarized in \cite{mukhtar2015vehicle}, most of collision detection schemes can be structured into four stages: image pre-processing, cueing stage, verification stage and tracking. The cueing stage aims to segment objects and scenes, and then identify the potential objects as regions-of-interest (ROI), where motion and appearance features are usually used to perform this process. Taking vehicle detection as an example, motion-based features are mainly generated by computing optical flow field from a moving object \cite{smith1995asset,cao2008vehicle,kuo2011vision}, and appearance features derive from some specific appearance of a vehicle such as shadow \cite{cheon2012vision, van2005vehicle}, edges \cite{betke2000real}, symmetry \cite{liu2007rear,kuehnle1991symmetry} and so on. After that, ROIs are processed further for determining whether they are vehicles at the verification stage, in which template matching \cite{bensrhair2001cooperative,lin2006multiple,hu2000tracking} and learning-based object classifier \cite{rybski2010visual,papageorgiou2000trainable,sun2002road} are exploited to recognize different objects. Final stage is tracking that takes advantage of temporal continuity to improve detection performance by using conventional Bayesian-filtering tracking techniques (i.e. Kalman filter \cite{moqqaddem2011spectral}, Extended Kalman filter \cite{barth2009estimating} and Particle filters \cite{chan2012vehicle}). The aforementioned vision-based detection techniques, however, are usually less accurate in highly dynamic complex environments. Meanwhile, these techniques rely on specific visual sensors (i.e. RGB-D, event-driven camera) or is directed at some specific objects. Recently, deep learning-based vehicle detection technology has achieved amazing success, which has a strong ability to solve classification and positioning problems at the same time \cite{song2019vision}. Such methods are typically divided into two categories: the two-stage method involving the aforementioned cueing and verification process (i.e. RCNN \cite{girshick2014rich}, Fast RCNN \cite{girshick2015fast}, Faster RCNN \cite{ren2015faster} and Mask RCNN \cite{he2017mask}), and the one-stage method integrating the above two processes into a unified optimization problem (i.e. SSD \cite{liu2016ssd} and YOLO framework \cite{redmon2016you,redmon2017yolo9000,redmon2018yolov3}). But most of deep learning-based methods are computationally expensive and have to run on GPUs for real-time execution. 

In fact, hundreds of millions of years of evolutionary development have enabled many animals on earth to detect approaching objects rapidly and robustly, which is mainly for escaping from predators, foraging and so forth. For instance, most insects have a special group of visual neurons sensitive to looming stimuli, such as the lobula plate tangential cells (LPTCs) in flies\cite{borst2014fly} and the giant movement detectors (LGMDs) in locusts \cite{gabbiani2002multiplicative,gabbiani2004multiplication,rind1996neural,rind2016two}. Inspired by neural mechanisms of their visual processing, some bioinspired models have emerged to simulate these collision detectors for improving the performance of collision detection and avoidance in robots. Serres and Ruffier \cite{serres2017optic}, for example, have reviewed a series of optical flow-based collision-free methods that are inspired by the sensitivity of LPTCs to optical flow. Such approaches are generally computational simplicity and successfully used in flying robots such as UAVs and MAVs \cite{bertrand2015bio,milde2015bioinspired,green2008optic,keshavan2015autonomous}. However, they are mainly suitable for lateral-collision avoidance in texture-rich environments \cite{green2008optic,serres2017optic} due to the constraint of optic flow balance hypothesis. Rotational motion and complex dynamic scenes have a serious impact on the performance since the quality of optical flow deteriorates or the balance hypothesis is destroyed. Based on optical flow, the study in \cite{humbert2010bioinspired} similarly proposed a wide-field method to extract visual cues for guidance and navigation efficiently. After that, a bioinspired approach was further introduced to extract the motion information about small-field objects \cite{escobaralvarez2019bioinspired,ohradzansky2018autonomous}, which was demonstrated computationally efficient and easier to implement in UAVs. But these approaches rely on the traditional optic flow computation and extract high-frequency content from planar optic flow. 

Another type of computational model is based on the LGMDs looming detectors\cite{yue2006collision,hu2016bio,fu2018shaping,fu2020improved}. These models, designed from neural circuits in locusts, are effective and robust to detect collision in a relatively stable environment. However, they can not distinguish radial and lateral motion well such that they are easily affected by translational or recessive motion, especially in cluttered and dynamic environments\cite{fu2020improved}. Fu et al. \cite{fu2020improved} attempted to solve this problem through splitting visual processing into ON and OFF channels, and using spike frequency adaptation. But all such models mediate the sensitivity of the output node to approach instead of a pixel-wise manner, thereby difficult to further extract motion information of colliding objects.

Neuroscience studies have shown that both the insect optic lobe and the mammalian retina have common neural circuit designs for elementary motion detection \cite{borst2015common,clark2016parallel}. But the mammalian motion vision is generally of higher quality and sensitivity than the insect visual system, which enables most mammals to detect insects and escape from predators quickly and accurately. A famous mammalian vision-inspired model is HMAX \cite{riesenhuber1999hierarchical}, a feedforward architecture for object recognition. However, the computational model of collision detection based on mammalian elementary motion vision are relatively few. Some early studies used a motion energy model to achieve elementary motion perception \cite{adelson1985spatiotemporal,heeger1988optical}, where Gabor filters and a temporal band-pass structure were expoited for spatiotemporal filtering. Such approaches usually work in an offline manner or with time delay since the temporal filtering needs to store a period of history and do filtering in a moving manner. Another related bioinspired model is the ViSTARS (Visually-guided Steering, Tracking, Avoidance, and Route Selection) \cite{browning2009a,browning2009cortical}, explaining how primates segment objects and compute heading from optic flow and obstacle avoidance in response to visual inputs. This model is biologically plausible but with a large number of dynamic parameters such that it is not easy to find suitable parameters in unknown environments.

This paper focuses on improving the performance of collision detection in complex moving backgrounds. Inspired by the elementary motion vision in the mammalian retina, this paper proposes a bioinspired approach-sensitive neural network model (ASNN) that can not only detect collision accurately and robustly in cluttered and dynamic backgrounds but also extract more collision information, like position and direction, for guiding rapid decision making. More specifically, there are three main contributions:
\begin{itemize}
  \item [1)] 
  A direction-selective visual processing module is built based on a spatiotemporal energy model, where a new neurodynamics-based temporal filter and 2-D spatial Gabor filter are exploited to perform spatiotemporal filtering in real time. This module can estimate motion direction accurately via only two mutually perpendicular spatiotemporal filtering processes.
  \item [2)]
  A novel approach-sensitive visual processing module is proposed, which is characterized by a push-pull structure formed by ON and OFF pathways. It responds strongly to approaching motion while insensitivity to lateral motion, by using a new antagonistic center-surround spatial filter.
  \item [3)]
  A mechanism of attention, composed of approach-sensitive attention and directionally selective attention, is introduced to suppress the lateral motion of cluttered backgrounds. Especially, we propose a method of directionally selective inhibition that can improve the performance of collision detection in cluttered and highly translational backgrounds effectively.
\end{itemize}

The rest of this paper is organized as follows. Section \Rmnum{2} introduces the neural mechanisms of approach sensitivity in the mammalian retina. Section \Rmnum{3} provides a detailed description of approach-sensitive neural network for collision detection. Section \Rmnum{4} describes the experimental setup and discusses the results of collision detection in synthetic and real scenarios respectively. Section \Rmnum{5} concludes this paper.

\section{NEURAL CIRCUIT OF APPROACH SENSITIVITY AND DIRECTION SELECTIVITY IN THE MAMMALIAN RETINA}
\begin{figure}[!htbp]
    \centering
    \includegraphics[width=0.45\textwidth]{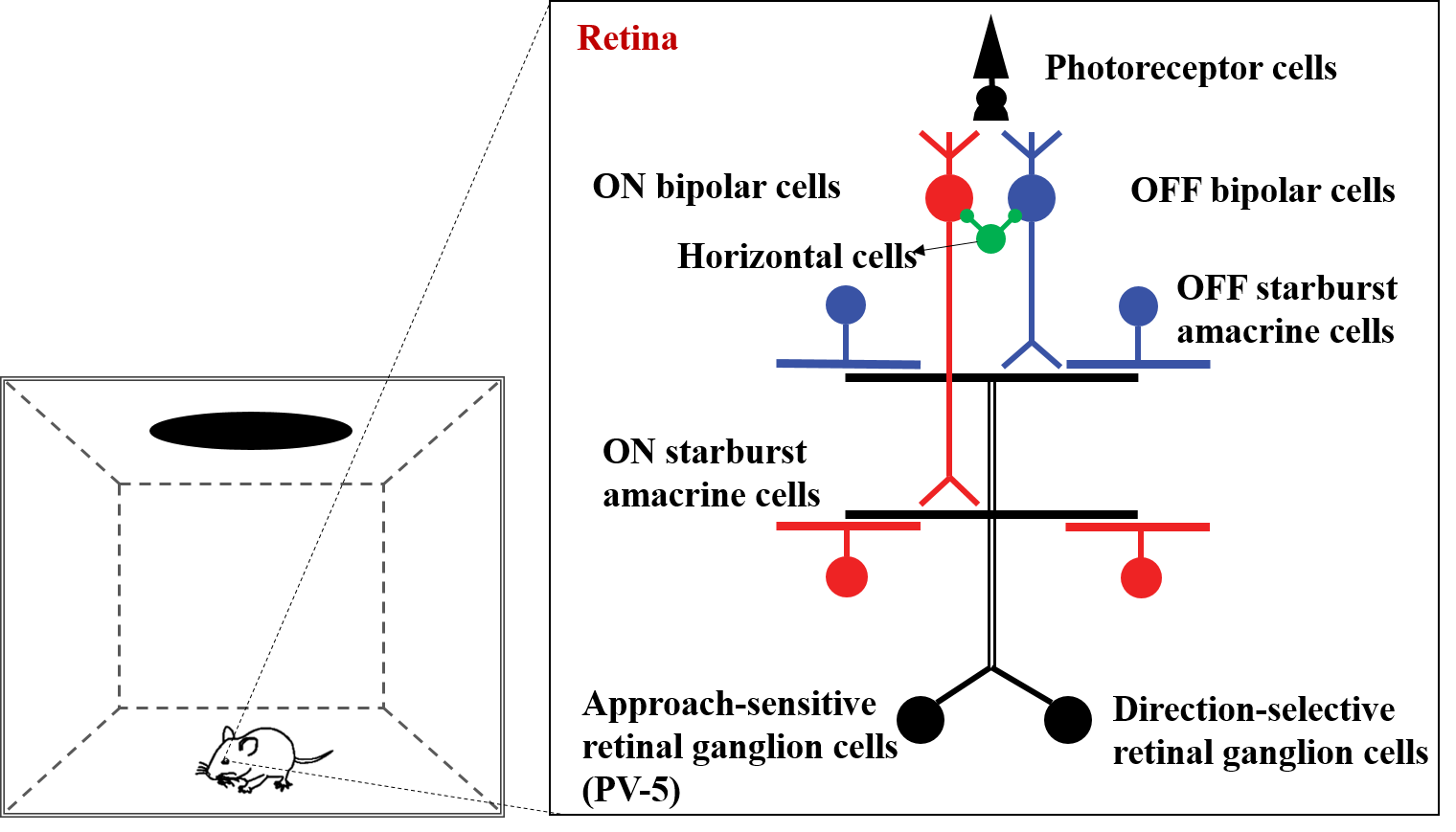}
    \caption{The neural circuits of elementary motion detection in the mouse retina.}
    \label{fig:s02f01}
\end{figure}

The mammalian retina is generally organized with a layered structure that consists of five principle cell types: photoreceptors, horizontal cells, bipolar cells, amacrine cells, and ganglion cells \cite{masland2001fundamental}, as shown in Fig.\ref{fig:s02f01}. Firstly, the photoreceptors, including rod and cone cells, are able to convert the optical signal into an electrical signal. The rod cells are mainly responsible for sensing the luminance of objects at night and the cone cells are sensitive to luminance and color in daylight. 

Then the photoreceptor signal is split and processed separately in parallel ON and OFF pathways: the ON channel responds to contrast increments and the OFF channel responds to contrast decrements \cite{clark2016parallel}. These two pathways are further integrated into different spatiotemporal dynamic processes in order to extract various elementary motion information. For example, the visual signals in two parallel channels are directly transmitted to the bipolar cells \cite{bialek1990temporal,burkhardt2007retinal}. In the bipolar cells, different synaptic types and connection of circuits can mediate the response speed of temporal dynamics such that fast and slow responses are produced. While different response speed of the bipolar cells are crucially important for direction selectivity and approach sensitivity in mammalian motion vision \cite{borst2015common,munch2009approach}. Meanwhile the bipolar cells are also affected by the inhibitory connections from the horizontal cells.

The visual signals are further transmitted to the amacrine cells, which are demonstrated to have the function of spatial filtering. Neuroscience studies have discovered a type of amacrine cell, named starburst amacrine cell (SAC), is sensitive to object motion in a specific direction. Each of the radial SAC dendrites responds to motion in its preferred direction, and converges to the SAC soma to form a specific preferred direction \cite{borst2015common}. Other directions are suppressed by the inhibitory effect of SAC dendrites. Additionally, neuroscientists have also uncovered an antagonistic center-surround receptive field organization in retinal On-SACs \cite{ankri2019antagonistic}. Repetitive visual stimulation can eliminate center and boost surround response in SAC, which is considered to be important for direction selectivity and approach sensitivity. 

Finally, retinal ganglion cells, as sole output neurons of the retina, can integrate the visual signals from the amacrine cells and the bipolar cells to obtain primary motion information, and communicate them to visual processing centers in the brain further. Neuroscience studies have discovered many different types of motion-sensitive ganglion cells so far, which contribute to three basic cases of motion sensitivity, including direction selectivity \cite{huberman2009genetic,kay2011retinal,trenholm2011parallel}, differential-motion sensitivity \cite{olveczky2003segregation} and approach sensitivity \cite{munch2009approach}. In what follows, we focus on the neural circuit of approach sensitivity.

In 2009, the study in \cite{munch2009approach} identified an approach-sensitive ganglion cell in the mouse retina called PV-5 cell that responds strongly to the approaching motion of dark objects instead of lateral and shrinking motion. This phenomenon derives from a rapid inhibitory pathway. The PV-5 cell receives excitatory and inhibitory inputs from bipolar cell and bipolar-amacrine circuit respectively. The excitatory signals are in OFF channel and mediated by chemical synapses such that the transmission is relatively slow. While the inhibitory signals are in ON channel and mediated by electrical synapses that make the transmission rapid. These two pathways form a push-pull structure, which can explain the sensitivity to approaching motion and the insensitivity to lateral motion well. In the lateral motion, the trailing (ON) edges may inhibit the excitatory effect of leading (OFF) edges. Whereas when dark object approaches, the ON signals keep silent, which makes the OFF signals burst without inhibition. Next, the part of approach-sensitive ganglion cells project to the superior colliculus, which constitutes a key neural circuit for detecting looming objects and triggering innate fear-escape responses \cite{shang2015a,shang2018divergent}.

\section{APPROACH-SENSITIVE NEURAL NETWORK FOR COLLISION DETECTION}
\begin{figure*}[!htbp]
    \centering
    \includegraphics[width=0.8\textwidth]{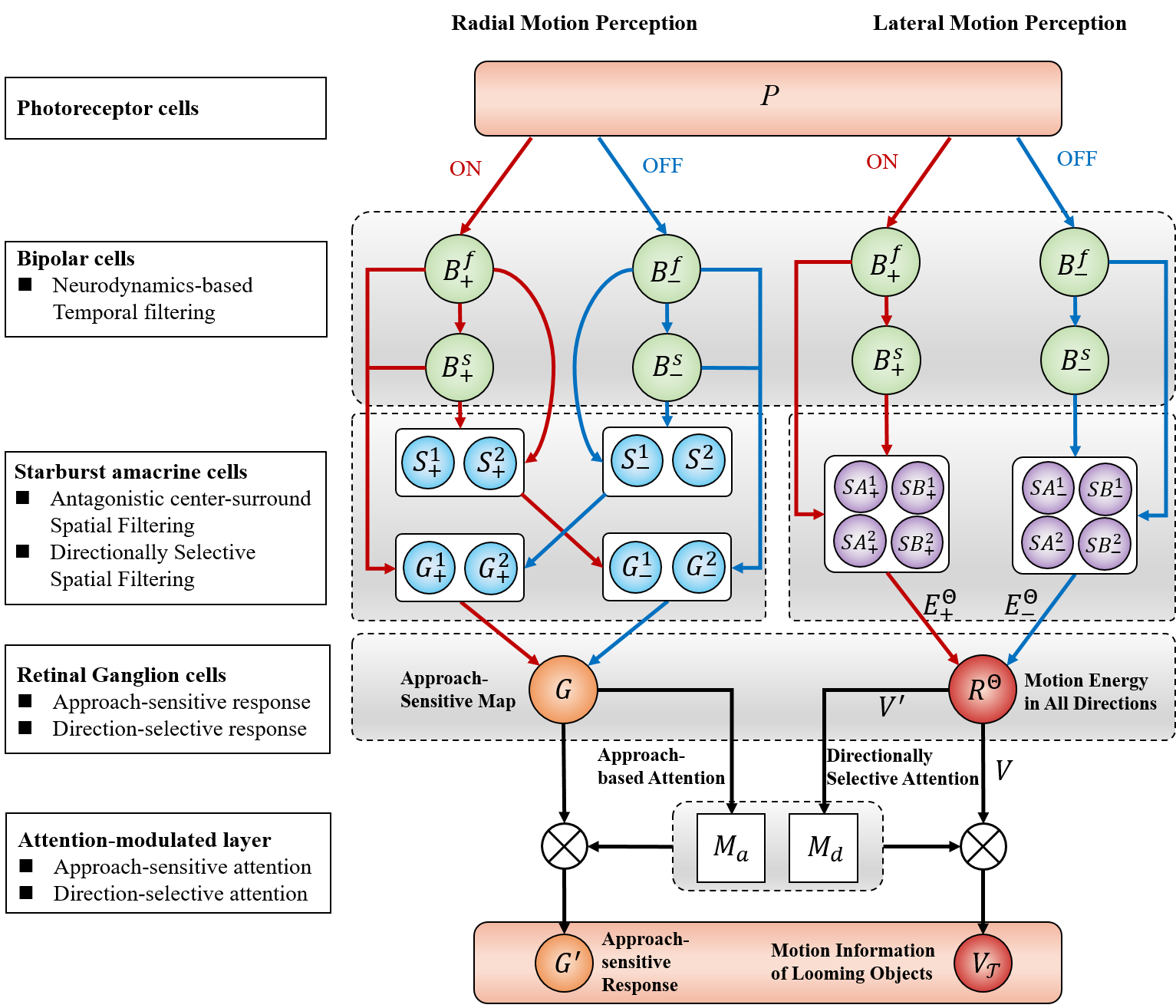}
    \caption{The computational neural network model of approach sensitivity.}
    \label{fig:s03f01}
\end{figure*}

Based on the neural mechanisms of approach sensitivity in the mammalian retina, an approach-selective neural network is proposed in this section. Specifically, this model consists of five layers: (1) Photoreceptor cells sense the change of luminance. (2) Bipolar cells split the visual signals into ON and OFF channels, and perform temporal filtering to produce fast and slow neural responses. (3) SACs perform spatial filtering in two fashions: antagonistic center-surround spatial filtering for radial motion perception and directionally selective spatial filtering for lateral motion perception. (4) Retinal ganglion cells compute approach-sensitive map and motion energy based on different spatiotemporal filtering information. (5) More collision information is extracted based on the results of radial and lateral motion perception. The computational structure is shown in Fig.\ref{fig:s03f01}.

\subsection{Photoreceptor cells}
The photoreceptor cells are mainly responsible for sensing the change of luminance. Assume the visual input is a gray-scale image without color information, these cells are then arranged as a 2D matrix that is the same size as the input image $I\in \mathbb{R}^{w\times h}$, where $w$ and $h$ are width and height of the image respectively. The activation of photoreceptor at time $t$ is defined the difference between adjacent frames.
\begin{equation}
P(x,y,t)=I(x,y,t)-I(x,y,t-1)+\sum_i^{N_p}p_iP(x,y,t-i),
\end{equation}
where $x$ and $y$ are the abscissa and ordinate of the input image respectively. $I(x,y,t)$ and $I(x,y,t-1)$ are the gray-scale value of two successive frames. Assume the change in luminance lasts for $N_p$ frames and decays with coefficient $p_i$. We define $p_i=(1+e^{ui})^{-1}$, and choose $u=1$.

\subsection{Bipolar cells}
The bipolar cells firstly split the visual signal into parallel ON and OFF channels. The ON channel responds to contrast increments and the OFF channel responds to contrast decrements, which can be modeled as a 'half-wave rectification'.
\begin{equation}
\begin{split}
B_{+}(x,y,t)&=\frac{1}{2}\left[ |P(x,y,t)|+P(x,y,t)\right],\\
B_{-}(x,y,t)&=\frac{1}{2}\left[ |P(x,y,t)|-P(x,y,t)\right].
\end{split}
\end{equation}

After that, difference of gaussians kernel is used to perform bandpass filtering for extracting regions of interest, which is based on the inhibitory effect of the horizontal cells and the amacrine cells. The kernel function is defined as follows.
\begin{equation}
g_I(x,y;\sigma_1,\sigma_2)=\left(\frac{F}{\sqrt{2\pi}\sigma_1}e^{\frac{-(x^2+y^2)}{2\sigma_1^2}}-\frac{F}{\sqrt{2\pi}\sigma_2}e^{\frac{-(x^2+y^2)}{2\sigma_2^2}}\right),
\end{equation}
where $F$ is the gain coefficient, $\sigma_1$ and $\sigma_2$ are the standard deviation of two gaussians. The responses of the ON and OFF bipolar cells can be computed through convolving with this kernel function.
\begin{equation}
\begin{split}
B_{+}^0(x,y,t)&=B_{+}(x,y,t)\otimes g_I,\\
B_{-}^0(x,y,t)&=B_{-}(x,y,t)\otimes g_I,
\end{split}
\end{equation}
where $B_{+}^0(x,y,t)$ and $B_{-}^0(x,y,t)$ are the neural activities in the first layer. $\otimes$ denotes the convolution operation.

Then a neurodynamics-based temporal filtering method is proposed to compute neural activities with different response speed. The structure of this filter is a n-level cascade of leaky integrators, denoted as $\mathcal{L}(K,n,z_0)$, where $K$ is the gain of filter, $n$ is the number of layers, and $z_0$ is the input. The dynamic equation of $n$th level neuron is defined as follows.
\begin{equation}\label{eqns3h5}
\tau \frac{\mathrm{d}z_n}{\mathrm{d}t}=-Az_n+Cz_{n-1},
\end{equation}
where $\tau$ is a time constant of integration, $A$ is a decay coefficient, and $C$ is a coefficient of transmission. The output of temporal filter is the difference between two layers of neurons. 
\begin{equation}\label{eqns3h6}
\mathcal{L}(K,n,z_0)=K(z_n-z_{n+1}),
\end{equation}

Then, the fast and slow responses of ON and OFF bipolar cells are computed respectively as follows.
\begin{equation}
\begin{split}
B_{+}^f(x,y,t)&=\mathcal{L}(K, n_f, B_{+}^0),\\
B_{+}^s(x,y,t)&=\mathcal{L}(K, n_s, B_{+}^0),\\
B_{-}^f(x,y,t)&=\mathcal{L}(K, n_f, B_{-}^0),\\
B_{-}^s(x,y,t)&=\mathcal{L}(K, n_s, B_{-}^0),
\end{split}
\end{equation}
where $n_f<n_s$.

This neurodynamics-based temporal filter serves as plausible approximations to psychophysical data, similar to the traditional temporal-filtering kernel described in \cite{adelson1985spatiotemporal}.
\begin{equation}
f(t)=\frac{(kt)^n}{n!}e^{-kt}-\frac{(kt)^{n+2}}{(n+2)!}e^{-kt}.
\end{equation}
However, the traditional temporal filtering takes an offline fashion that convolves or correlates the signal history with above kernel function. If the window size of filter is relatively large or the dimension of input image is very high, the problem of time delay will arise such that the real-time performance will be severely affected. While the proposed filter performs the temporal filtering only through reading out the activity of Leaky neurons directly without any time delay. It can precisely produce temporal sequences at different response speeds, which is crucially important for perceiving radial and lateral motion. More analysis of this temporal filter is described in the Appendix.

\begin{figure}[!htbp]
    \centering
    \subfigure[]{\label{fig:s03f02a}\includegraphics[width=0.4\textwidth]{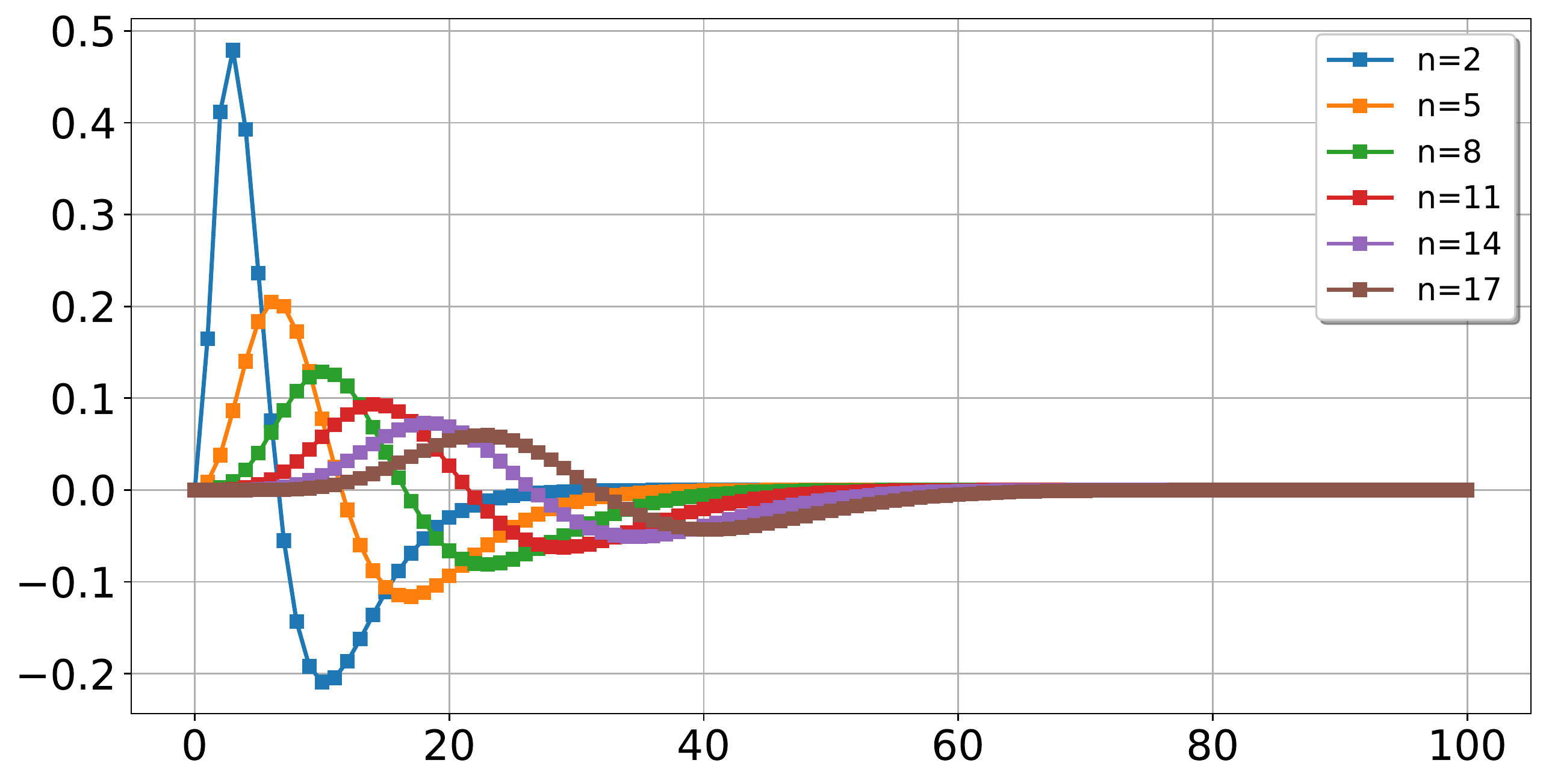}}\\
    \subfigure[]{\label{fig:s03f02b}\includegraphics[width=0.4\textwidth]{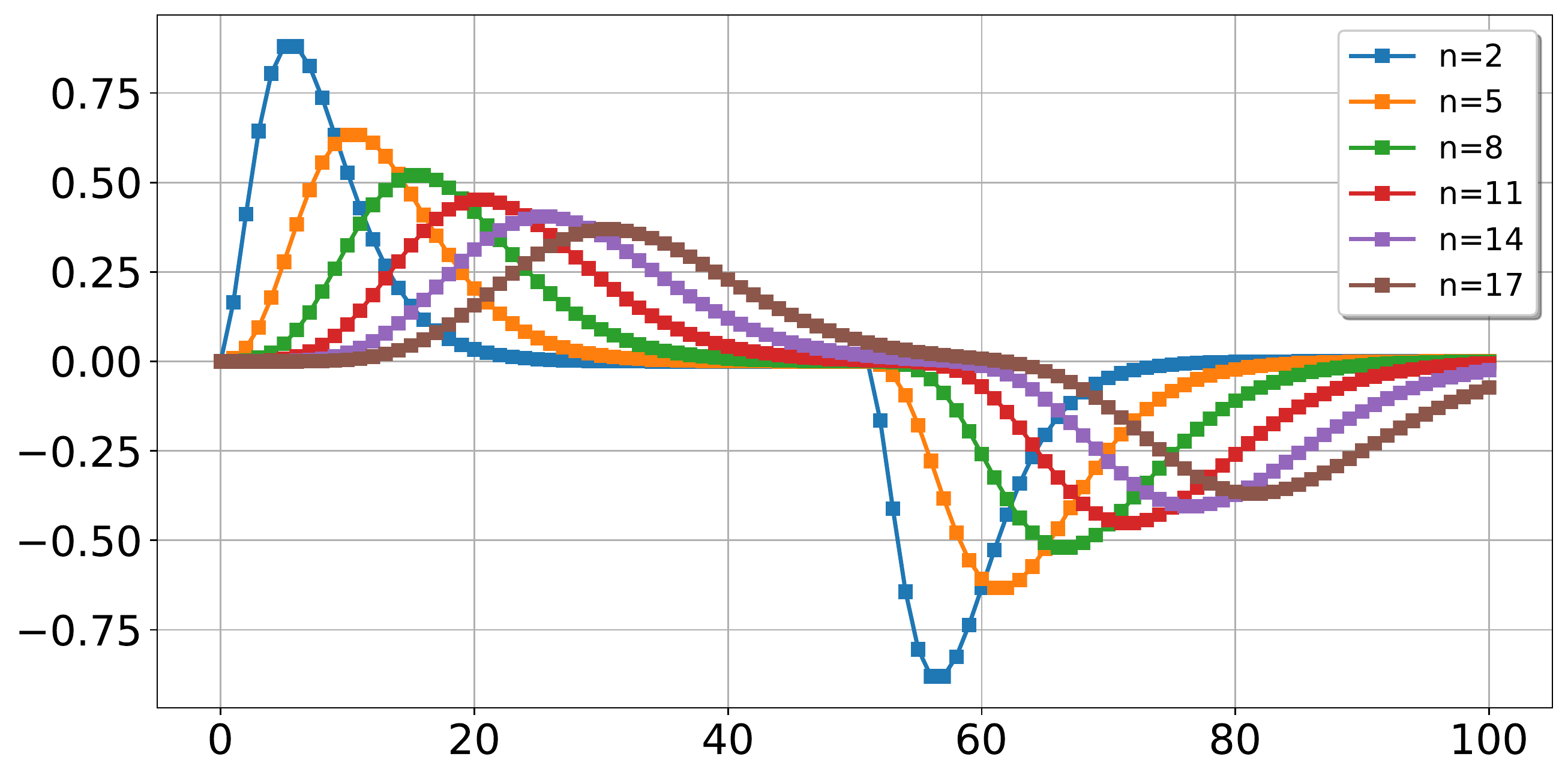}}
    \caption{(a) The unit impulse response of temporal filter at different $n$. (b) The step response of temporal filter at different $n$.}
    \label{fig:s03f02}
\end{figure}

A simple simulation is conducted further to illustrate the characteristics of this filter. The parameters are set to $K=5$, $C=60$ and $\tau=8$. The simulation time is $100$ steps, and step size is $\Delta t =0.05s$. The unit impulse response and step response of the temporal filter at different $n$ are shown in Fig.\ref{fig:s03f02a} and Fig.\ref{fig:s03f02b} respectively. Obversely, the response speed decreases with the increase of $n$. Meanwhile, the response amplitude is also related to the number of cascades $n$. As $n$ increases, the response gradually weakens.

\subsection{Amacrine cells}
\subsubsection{Directionally selective spatial filtering for lateral motion perception} 
\begin{figure}[!htbp]
    \centering
    \includegraphics[width=0.4\textwidth]{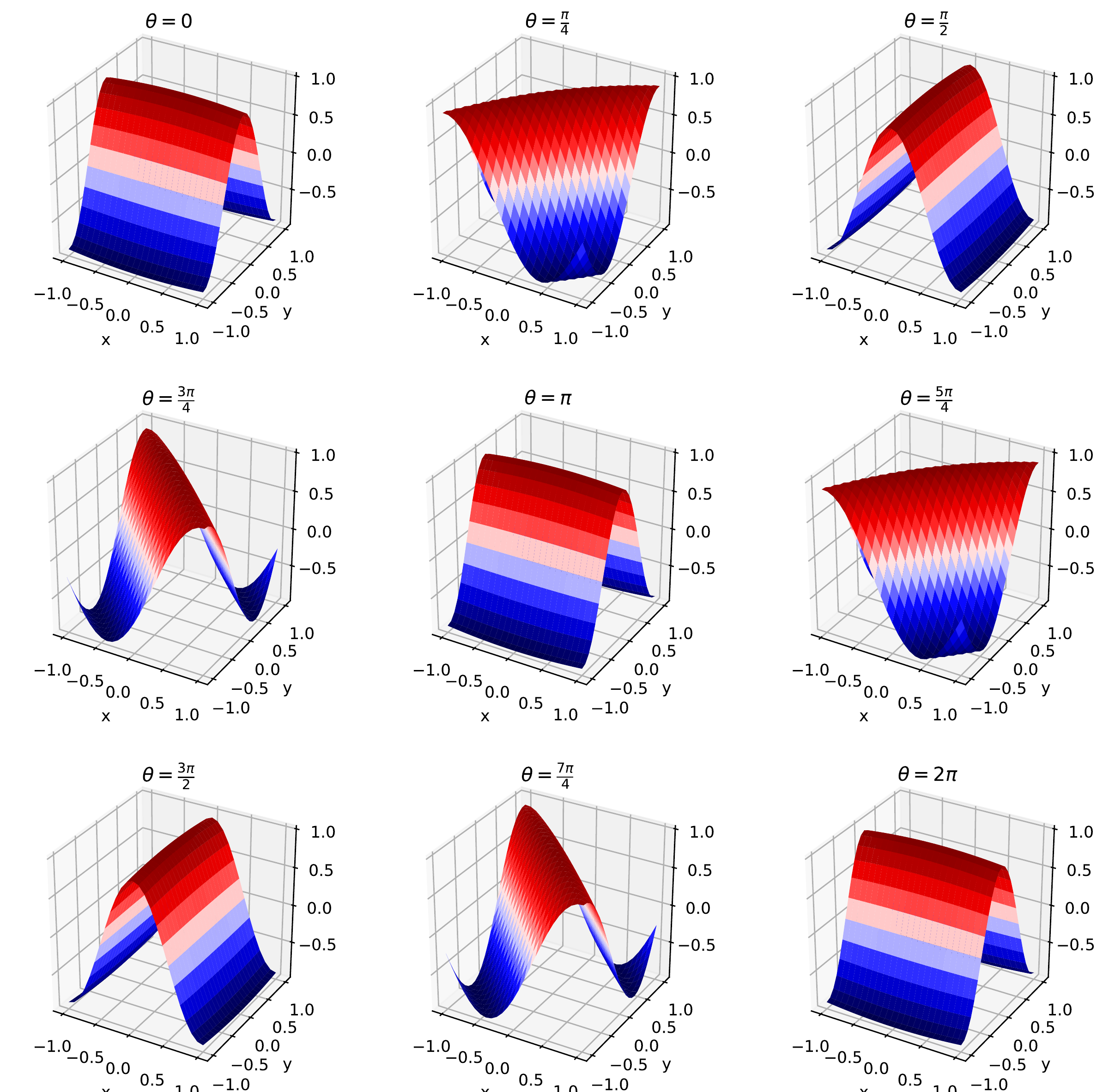}
    \caption{The shape of Gabor filters at different orientations.}
    \label{fig:s03f04}
\end{figure}

Accurately extracting motion information about colliding targets is essential for robots to make rapid decisions, in which direction perception is one of the most important factors. SACs can perform direction-selective spatial filtering. According to the findings of neuroscience, frequency and orientation representations of Gabor filters are demonstrated to be similar to those of the human visual system. Herein, Gabor filters are used to simulate the direction selectivity, where the kernel function is
\begin{equation}
g_{DS}(x,y)=\exp\left(-\frac{x^{\prime 2}+y^{\prime 2}}{2\sigma_{DS}^2}\right)\cos\left(\frac{2\pi x^{\prime}}{\lambda_{DS}}+\psi\right),
\end{equation}
where $x'=x\cos(\theta)+y\sin(\theta)$, $y'=-x\sin(\theta)+y\cos(\theta)$. $\lambda_{DS}$ is wavelength of the sinusoidal factor. $\theta$ represents the orientation of the Gabor function. $\psi_{DS}$ is the phase offset. $\sigma_{DS}$ is the standard deviation of the Gaussian envelope. The shape of Gabor kernel function at different orientations is shown in Fig.\ref{fig:s03f04}, where $\lambda_{DS}=2, \sigma_{DS}=3, \psi_{DS}=0$.

In order to compute the motion energy, two Gabor filters ($\psi_{DS}=0$ and $\psi_{DS}=\frac{\pi}{2}$) are respectively chosen to process the visual signals from the bipolar cells. Denote them as $g^{0}_{DS}$ and $g^{\frac{\pi}{2}}_{DS}$. At each orientation $\theta$, eight types of SAC outputs are calculated in the following way.
\begin{equation}
\begin{split}
	SA_{+}^1(x,y,t)&=B_{+}^s(x,y,t)\otimes g^{0}_{DS},\\
	SB_{+}^1(x,y,t)&=B_{+}^s(x,y,t)\otimes g^{\frac{\pi}{2}}_{DS},\\
	SA_{+}^2(x,y,t)&=B_{+}^f(x,y,t)\otimes g^{0}_{DS},\\
	SB_{+}^2(x,y,t)&=B_{+}^f(x,y,t)\otimes g^{\frac{\pi}{2}}_{DS},\\
	SA_{-}^1(x,y,t)&=B_{-}^s(x,y,t)\otimes g^{0}_{DS},\\
	SB_{-}^1(x,y,t)&=B_{-}^s(x,y,t)\otimes g^{\frac{\pi}{2}}_{DS},\\
	SA_{-}^2(x,y,t)&=B_{-}^f(x,y,t)\otimes g^{0}_{DS},\\
	SB_{-}^2(x,y,t)&=B_{-}^f(x,y,t)\otimes g^{\frac{\pi}{2}}_{DS}.
\end{split}
\end{equation}
These responses are fed into the ganglion cells further to compute the motion energy of lateral motion.

\subsubsection{Antagonistic center-surround spatial filtering for radial motion perception} 
\begin{figure}[!htbp]
    \centering
    \includegraphics[width=0.3\textwidth]{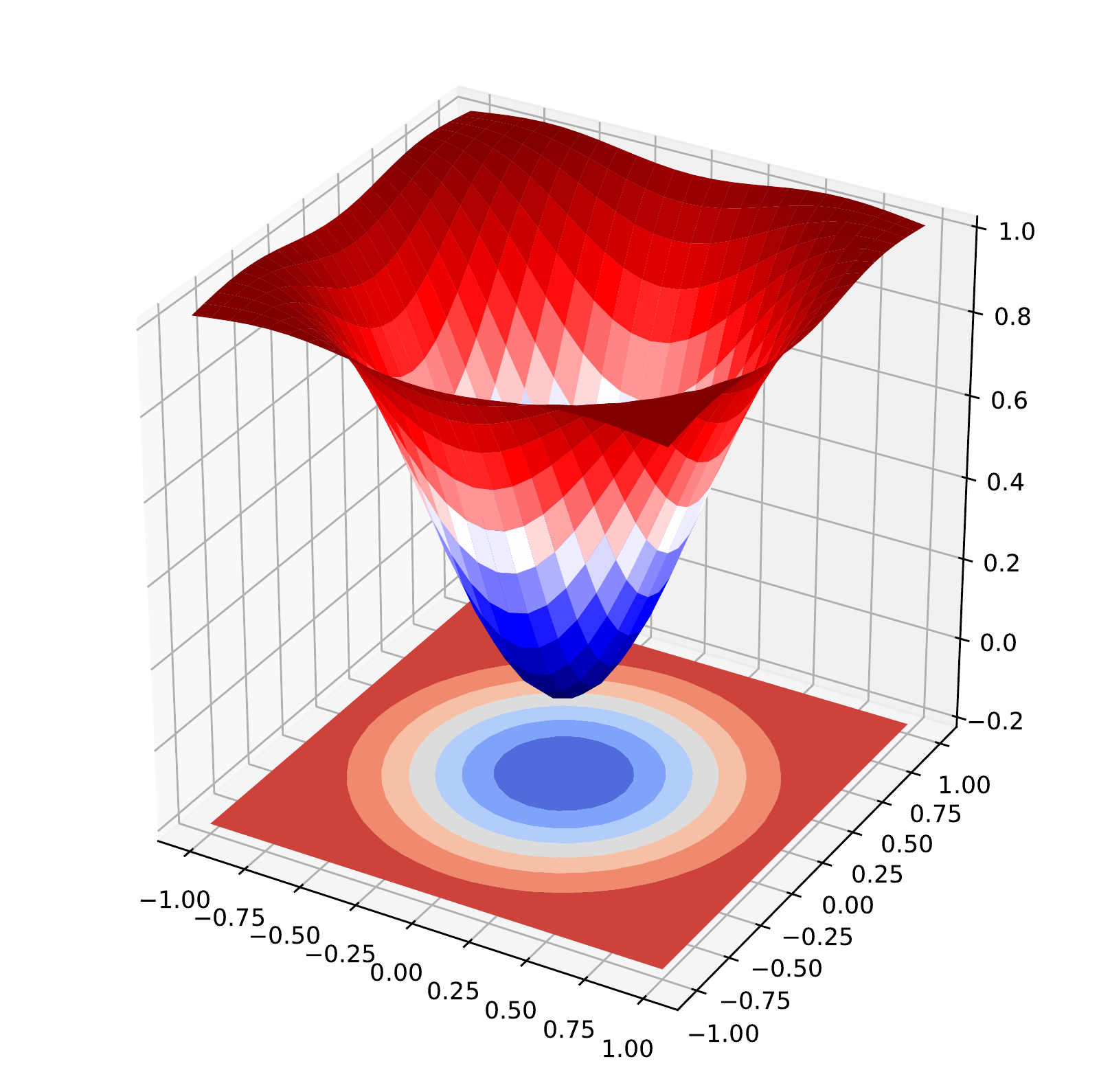}
    \caption{The shape of the antagonistic center-surround kernel function.}
    \label{fig:s03f05}
\end{figure}

As mentioned above, SACs are not only directionally selective, but also have an antagonistic center-surround receptive field found in recent study \cite{ankri2019antagonistic}. We consider that the directionally-selective spatial filtering may contribute to sense the direction of lateral motion, while the antagonistic center-surround spatial filtering is mainly related to looming stimuli in radial motion. Based on the formation of Gabor filters, we define a new kernel function of antagonistic center-surround spatial filtering as follows.
\begin{equation}
g_{CS}(x,y)=1-\exp\left(-\frac{x^2+y^2}{2\sigma_{CS}^2}\right)\cos\left(\frac{2\pi (x^2+y^2)}{\lambda_{CS}}+\psi\right),
\end{equation}
where $\lambda_{CS}$ is wavelength of the sinusoidal factor, which is usually measured in pixels. $\psi_{CS}$ is the phase offset. $\sigma_{CS}$ is the standard deviation of the Gaussian envelope. The shape of center-surround kernel function is shown in Fig.\ref{fig:s03f05}, where $\lambda_{CS}=4, \sigma_{CS}=0.5, \psi_{CS}=0$. It reflects the phenomenon that repetitive visual stimulation can eliminate center and boost surround response in SACs.

The visual signals from ON-Bipolar cells arrive at the ON-SACs rapidly due to electrical synapses. Fast responses of ON-Bipolar cells are used to compute the inhibitory signals of ON-SACs through convolving the above kernel function.
\begin{equation}
S^1_{+}(x,y,t)=relu(B_{+}^f(x,y,t)\otimes g_{CS}),
\end{equation}
where $relu(x)=\max(0,x)$ is a rectified linear unit, $\otimes$ represents the convolution operation. This process can eliminate center and boost surround response in all directions, which makes the inhibition area overlap with the excitatory area of slow OFF pathway in lateral motion.

Currently, neuroscience studies support the approach sensitivity to dark objects, in other word this sensitivity is only OFF type. We consider it is also ON type that is sensitive to bright objects in a dark environment. In this pathway, fast responses of OFF-Bipolar cells are used to compute the inhibitory signals of OFF-SACs as follows.
\begin{equation}
S^1_{-}(x,y,t)=relu(B_{-}^f(x,y,t)\otimes g_{CS}).
\end{equation}
Likewise, this inhibition area overlaps with the excitatory area of slow ON pathway in lateral motion.

In addition, the lateral motion of an object generally can trigger both ON and OFF activations simultaneously. For instance, in a scene where a black block moves against a white background, the 'leading' edge can cause OFF response and the 'trailing' edge can cause ON response. The inhibitory signals defined above are only able to suppress the 'leading' response rather than the 'trailing' response in lateral motion. We assume that the slow 'leading' response can also have inhibitory effect on the 'trailing' response through the antagonistic center-surround spatial filtering, especially when the moving object is relatively small. These inhibitory signals in ON and OFF pathways are formalized as follows. 
This process can suppress the noise of lateral motion further.
\begin{equation}
\begin{split}
	S^2_{+}(x,y,t)&=relu(B_{+}^s(x,y,t)\otimes g_{CS}),\\
	S^2_{-}(x,y,t)&=relu(B_{-}^s(x,y,t)\otimes g_{CS}).
\end{split}
\end{equation}

\subsection{Retinal ganglion cells}
\subsubsection{Motion Energy of Lateral Motion}
As mentioned in the section \Rmnum{2}, many types of ganglion cells are sensitive to different motion patterns, including direction selectivity and approach sensitivity. For direction selectivity, some ganglion cells are able to fuse the spatiotemporal filtering results to calculate the motion energy of each pixel such that the motion direction can be estimated further. In this paper, direction perception is used to perform directionally selective inhibition for suppressing background motion and determine the motion direction of colliding targets, which is important for animals and robots to make rapid and correct decisions. Specifically, the responses of ganglion cells at filtering orientation $\theta$ can be obtained as follows.
\begin{equation}
\begin{split}
E^{\theta}_{+}(x,y,t)&=SA_{+}^1(x,y,t)\cdot SB_{+}^2(x,y,t)\\
&\quad -SA_{+}^2(x,y,t)\cdot SB_{+}^1(x,y,t),\\
E^{\theta}_{-}(x,y,t)&=SA_{-}^1(x,y,t)\cdot SB_{-}^2(x,y,t)\\
&\quad-SA_{-}^2(x,y,t)\cdot SB_{-}^1(x,y,t).
\end{split}
\end{equation}

One type of the ganglion cells is known as ON-OFF direction-selective retinal ganglion cell that responds to its preferred direction in both ON and OFF pathways. Based on this, we combine the motion energy of two parallel pathways as follows.
\begin{equation}
R^{\theta}(x,y,t)=v_{+}E^{\theta}_{+}(x,y,t)+v_{-}E^{\theta}_{-}(x,y,t),
\end{equation}
Denote $\mathbf{R}^{\Theta}$ as the set of neural outputs in all filtering orientations $\Theta$. We obtain the motion energy map of lateral motion in all directions through a maximization operation, which is formalized as follows.  
\begin{equation}
V(x,y,t)=\max_{\theta}(\mathbf{R}^{\Theta}(x,y,t)).
\end{equation}
Importantly, the motion direction $\varphi$ can be estimated via only two mutually perpendicular spatial filtering processes.
\begin{equation}
\hat{\varphi}(x,y,t) = \arctan2(R^{\frac{\pi}{2}}(x,y,t), R^{0}(x,y,t)).
\end{equation}

\subsubsection{Approach-sensitive Map of Radial Motion}
Retinal ganglion cells combine excitatory and inhibitory inputs from slow OFF-Bipolar cells and fast ON-SACs respectively to compute approach-sensitive responses of OFF type $G^1_{-}$. They also integrate slow excitatory and fast inhibitory inputs from ON-Bipolar cells and OFF-SACs to obtain ON-type ones $G^1_{+}$. Meanwhile we introduce the 'leading' responses $S^2_{+}$ and $S^2_{-}$ to inhibit the 'trailing' responses $B_{-}^f$ and $B_{+}^f$ respectively in lateral motion. Formally, four kinds of retinal ganglion cells are defined as follows.
\begin{equation}
\begin{split}
G^1_{+}(x,y,t)&=relu\left[relu(B_{+}^s(x,y,t))-S^1_{-}(x,y,t) \right ],\\
G^2_{+}(x,y,t)&=relu\left[relu(B_{+}^f(x,y,t))-S^2_{-}(x,y,t) \right ],\\
G^1_{-}(x,y,t)&=relu\left[relu(B_{-}^s(x,y,t))-S^1_{+}(x,y,t) \right ],\\
G^2_{-}(x,y,t)&=relu\left[ relu(B_{-}^f(x,y,t))-S^2_{+}(x,y,t) \right].\\
\end{split}
\end{equation}
Then the signals of ON and OFF channels can be fused by weighted average as follows.
\begin{equation}
\begin{split}
G(x,y,t)&=w^1_{+}G^1_{+}(x,y,t)-w^2_{-}G^2_{-}(x,y,t)\\
&\quad +w^1_{-}G^1_{-}(x,y,t)-w^2_{+}G^2_{+}(x,y,t).
\end{split}
\end{equation}
where $w^1_{+}$, $w^2_{-}$, $w^1_{-}$ and $w^2_{+}$ are weighting coefficients.

We further introduce a attention-modulated layer that aims to suppress the lateral motion of cluttered backgrounds and locate potential areas of looming objects. In these areas, more motion information about collision is extracted for rapid and accurate decision making. More specifically, this attention is composed of two parts. The first one is named as approach-sensitive attention that is produced through Gaussian blur on the approach-sensitive map. The kernel function is a two-dimensional Gaussian function.
\begin{equation}
g_{\sigma}(x,y)=\frac{1}{2\pi\sigma_g^2}\exp\left(-\frac{x^2+y^2}{2\sigma_g^2}\right).
\end{equation}
Large pixel radius and standard deviation are generally chosen to smooth the approach-sensitive image through the following convolution course.
\begin{equation}
G_{\sigma}(x,y,t)=G(x,y,t)\otimes g_{\sigma}.
\end{equation}
Next in order to suppress background noise, some small isolated excitations are eliminated by binary thresholding. The attention mask is generated as follows. 
\begin{equation}
M_a(x,y,t)=\mathcal{B}(G_{\sigma}(x,y,t),\gamma_a),
\end{equation}
where $\mathcal{B}$ is a binarization function with a threshold value $\gamma_a$. This process marks potentially looming targets as 1, and others as 0.

The second part is called as directionally selective attention. In a real scenario, we found that the collision detection of robotic visual system is seriously affected by the lateral motion of cluttered backgrounds. Fast-moving background can activate many approach-sensitive ganglion cells, where the 'trailing' responses can not inhibit the 'leading' response effectively due to faster movement speed and a larger moving region. However, one of the characteristics of background lateral motion is that many moving areas have the same direction of movement. Thus we propose a method of directionally selective inhibition in which top-$k$ directions with the largest motion energy are inhibited. Through this operation, most moving backgrounds will be suppressed in the case of lateral motion. While for radial motion, motion energies in all directions generally increase simultaneously such that the top-$k$ inhibition only partially weakened the activation of ganglion cells. Formally, we record the sum of motion energy in each filtering orientation $\theta$ as follows.
\begin{equation}
R_{sum}(\theta,t)=\sum_{x,y}R^{\theta}(x,y,t).
\end{equation}
Then top-$k$ directions with the largest sum of motion energy are added to the set of inhibitory direction $\Theta_{I}$. The inhibited activation is computed in the following way.
\begin{equation}
V^{\prime}(x,y,t) = V(x,y,t)-\max_{\theta}\mathbf{R}^{\Theta_I}.
\end{equation}
Similarly, the directionally selective attention mask is also obtained by binary thresholding.
\begin{equation}
M_d(x,y,t)=\mathcal{B}(V^{\prime}(x,y,t)\otimes g_{\sigma},\gamma_d).
\end{equation}

Finally the activation of retinal ganglion cells is modified through multiplying by the attention mask.
\begin{equation}
G^{\prime}(x,y,t)=G(x,y,t)\cdot M_a(x,y,t)\cdot M_d(x,y,t).
\end{equation}
The collision is detected based on sum of the approach-sensitive responses.
\begin{equation}
u(t)=\sum_{x,y} G^{\prime}(x,y,t).
\end{equation}

\subsection{Collision Information Extraction}
When collision is detected, rapid decisions also rely on more accurate information, including position, direction, velocity of the target and so on. Herein, the region of attention will be recognized as the region of the colliding targets if a collision event is detected. In other words, if $M_a(x,y,t)\cdot M_d(x,y,t)=1$, the location $(x,y)$ will be added to the regions of interest. Then we use clustering algorithms, such as DBSCAN \cite{ester1996density,schubert2017dbscan}, to discover the clusters of looming targets and to determine the position of them. The clustering center $(X,Y)$ is regarded as the position of the candidate target. Meanwhile, the direction and motion energy of them are estimated through a population coded algorithm. Specifically, the motion energy of each point can be decomposed into two parts along the $x$ and $y$ axis.
\begin{equation}
\begin{split}
V_x(x,y,t) &= V(x,y,t)\cos(\hat{\varphi}(x,y,t)),\\
V_y(x,y,t) &= V(x,y,t)\sin(\hat{\varphi}(x,y,t)).
\end{split}
\end{equation}
The motion energy of each cluster is estimated via the mean value method.
\begin{equation}
\begin{split}
V_X(x,y,t) &= \frac{1}{N_{\mathcal{T}}}\sum_{(x,y)\in \mathcal{T}} V_x(x,y,t),\\
V_Y(x,y,t) &= \frac{1}{N_{\mathcal{T}}}\sum_{(x,y)\in \mathcal{T}} V_y(x,y,t),
\end{split}
\end{equation}
where $\mathcal{T}$ represents the point set of a given target, $N_{\mathcal{T}}$ is the number of points in this target. The direction of the candidate looming target is further computed as follows.
\begin{equation}
\begin{split}
\varPhi(X,Y,t) &= \arctan2\left(V_X(x,y,t), V_Y(x,y,t)\right),\\
V(X,Y,t) &= \sqrt{V_X(x,y,t)^2+V_Y(x,y,t)^2}.
\end{split}
\end{equation}

Algorithm \ref{alg_1} presents the pseudocode for estimating motion directions. Algorithm \ref{alg_2} presents the pseudocode for the computational model of approach sensitivity. 
\IncMargin{1em}
\begin{algorithm}[!htbp]
\label{alg_1}
\caption{The computational process of directional selectivity.}
\SetKwFunction{ComputePhotoreceptor}{ComputePhotoreceptor}
\SetKwFunction{ComputeFirstBipolar}{ComputeFirstBipolar}
\SetKwFunction{TemporalFiltering}{TemporalFiltering}
\SetKwFunction{SpatialFiltering}{SpatialFiltering}
\SetKwFunction{ComputeGanglion}{ComputeGanglion}
\SetKwFunction{DirectionEstimation}{DirectionEstimation}
\SetKwInOut{Input}{input}\SetKwInOut{Output}{output}
\KwIn{The grayscale image of current frame.}
\KwOut{The estimated direction $\hat{\varphi}$ of each pixel.}
\BlankLine
Initialize the model parameters, and record the initial image $I_1$\;
\For{$t \leftarrow 2$ \KwTo $T$}{
Store the previous image $I_{t-1}$, and get the current image $I_t$\;
$P_t\leftarrow $ \ComputePhotoreceptor{$I_t$,$I_{t-1}$}\;
$B_{+}^0,B_{-}^0\leftarrow$ \ComputeFirstBipolar{$P_t$,$g_I$}\;
\tcp{\emph{Perform temporal filtering to obtain responses of the bipolar cells.}}
$B_{+}^f,B_{+}^s\leftarrow$ \TemporalFiltering{$B_{+}^0$}\;
$B_{-}^f,B_{-}^s\leftarrow$ \TemporalFiltering{$B_{-}^0$}\;
\For{$\theta \leftarrow 0$ \KwTo $2\pi$}{
\tcp{\emph{Perform directionally selective spatial filtering to compute responses of the amacrine cells.}}
$SA_{+}^1\leftarrow$ \SpatialFiltering{$B_{+}^s,g^0_{DS}$}\;
$SB_{+}^1\leftarrow$ \SpatialFiltering{$B_{+}^s,g^{\frac{\pi}{2}}_{DS}$}\;
$SA_{+}^2\leftarrow$ \SpatialFiltering{$B_{+}^f,g^0_{DS}$}\;
$SB_{+}^2\leftarrow$ \SpatialFiltering{$B_{+}^f,g^{\frac{\pi}{2}}_{DS}$}\;
$SA_{-}^1\leftarrow$ \SpatialFiltering{$B_{-}^s,g^0_{DS}$}\;
$SB_{-}^1\leftarrow$ \SpatialFiltering{$B_{-}^s,g^{\frac{\pi}{2}}_{DS}$}\;
$SA_{-}^2\leftarrow$ \SpatialFiltering{$B_{-}^f,g^0_{DS}$}\;
$SB_{-}^2\leftarrow$ \SpatialFiltering{$B_{-}^f,g^{\frac{\pi}{2}}_{DS}$}\;
\tcp{\emph{Compute responses of the ganglion cells based on the motion energy method.}}
$E_{+}^{\theta},E_{-}^{\theta},R_l^{\theta}\leftarrow$ \ComputeGanglion{$SA_{+}^1,SB_{+}^1,SA_{+}^2,SB_{+}^2$}\;
$\theta\leftarrow \theta + \frac{\pi}{4}$ 
}
$\hat{\varphi}\leftarrow$ \DirectionEstimation{$R_l^0,R_l^{\frac{\pi}{2}}$}\;
}
\end{algorithm}
\DecMargin{1em}

\IncMargin{1em}
\begin{algorithm}[!htbp]
\label{alg_2}
\caption{The computational process of approach selectivity.}
\SetKwFunction{ComputePhotoreceptor}{ComputePhotoreceptor}
\SetKwFunction{ComputeFirstBipolar}{ComputeFirstBipolar}
\SetKwFunction{TemporalFiltering}{TemporalFiltering}
\SetKwFunction{SpatialFiltering}{SpatialFiltering}
\SetKwFunction{ComputeGanglion}{ComputeGanglion}
\SetKwFunction{IntegrateGanglion}{IntegrateGanglion}
\SetKwFunction{ApproachAttention}{ApproachAttention}
\SetKwFunction{DirectionAttention}{DirectionAttention}
\SetKwInOut{Input}{input}\SetKwInOut{Output}{output}
\KwIn{The grayscale image of current frame.}
\KwOut{The approach-sensitive activities $G'$ of looming targets.}
\BlankLine
Initialize the model parameters, and record the initial image $I_1$\;
\For{$t \leftarrow 2$ \KwTo $T$}{
Store the previous image $I_{t-1}$, and get the current image $I_t$\;
$P_t\leftarrow $ \ComputePhotoreceptor{$I_t$,$I_{t-1}$}\;
$B_{+}^0,B_{-}^0\leftarrow$ \ComputeFirstBipolar{$P_t$,$g_I$}\;
\tcp{\emph{Perform temporal filtering to obtain responses of the bipolar cells.}}
$B_{+}^f,B_{+}^s\leftarrow$ \TemporalFiltering{$B_{+}^0$}\;
$B_{-}^f,B_{-}^s\leftarrow$ \TemporalFiltering{$B_{-}^0$}\;
\tcp{\emph{Perform antagonistic center-surround spatial filtering in the SACs.}}
$S^1_{+}\leftarrow$ \SpatialFiltering{$B_{+}^f,g_{CS}$}\;
$S^1_{-}\leftarrow$ \SpatialFiltering{$B_{-}^f,g_{CS}$}\;
$S^2_{+}\leftarrow$ \SpatialFiltering{$B_{+}^s,g_{CS}$}\;
$S^2_{-}\leftarrow$ \SpatialFiltering{$B_{-}^s,g_{CS}$}\;
\tcp{\emph{Combine excitatory and inhibitory inputs to obtain responses of the retinal ganglion cells.}}
$G^1_{+}\leftarrow$ \ComputeGanglion{$B_{+}^s,S^1_{-}$}\;
$G^2_{+}\leftarrow$ \ComputeGanglion{$B_{+}^f,S^2_{-}$}\;
$G^1_{-}\leftarrow$ \ComputeGanglion{$B_{-}^s,S^1_{+}$}\;
$G^2_{-}\leftarrow$ \ComputeGanglion{$B_{-}^f,S^2_{+}$}\;
\tcp{\emph{Integrate the signals of ON and OFF channels to obtain the approach-sensitive map.}}
$G\leftarrow$ \IntegrateGanglion{$G^1_{+},G^2_{+},G^1_{-},G^2_{-}$}\;
}
\tcp{\emph{Compute the mask of approach-sensitive attention.}}
$M_a\leftarrow$ \ApproachAttention{$G_{\sigma},g_{\sigma}$}\;
\tcp{\emph{Compute the mask of directionally selective attention.}}
$M_d\leftarrow$ \DirectionAttention{$\mathbf{R}^{\Theta},\mathbf{R}^{\Theta_I}$}\;
$G'=G \cdot M_a \cdot M_d$\;

\end{algorithm}
\DecMargin{1em}

\section{EXPERIMENTAL EVALUATION}
\subsection{Experimental Setup}
The synthetic experiments and vehicle collision implementation have been validated in a workstation. The workstation has an AMD Ryzen 5 3600X 6-Core 3.80GHz Processor and 16GB of RAM. It runs an Ubuntu 18.04 LTS 64-b operating system. The architecture has been developed in Python 3.8. The real robotic platform is a tracked robot named Komodo-02, as shown in Fig.\ref{fig:s04f09}. The architecture has been developed in C++/Python under the ROS Melodic framework.

Following three groups of experiments are conducted to verify the the accuracy and effectiveness of the proposed algorithm. The first one is to test the basic properties of our proposed framework in several synthetic scenes, comparing with previous computational models. Next, we explore the potential of the proposed collision perception systems in the ground vehicle applications, where on-road recordings from vehicle camera recorders are fed into the ASNN model. Finally, the proposed model is implemented in our real robotic platform.

\subsection{Approach Sensitivity in Synthetic Scenarios}
\begin{figure*}[!htbp]
    \centering
    \subfigure[]{
    	\label{fig:s04f01a}
		\begin{minipage}[b]{0.3\linewidth}
		\includegraphics[width=1\linewidth]{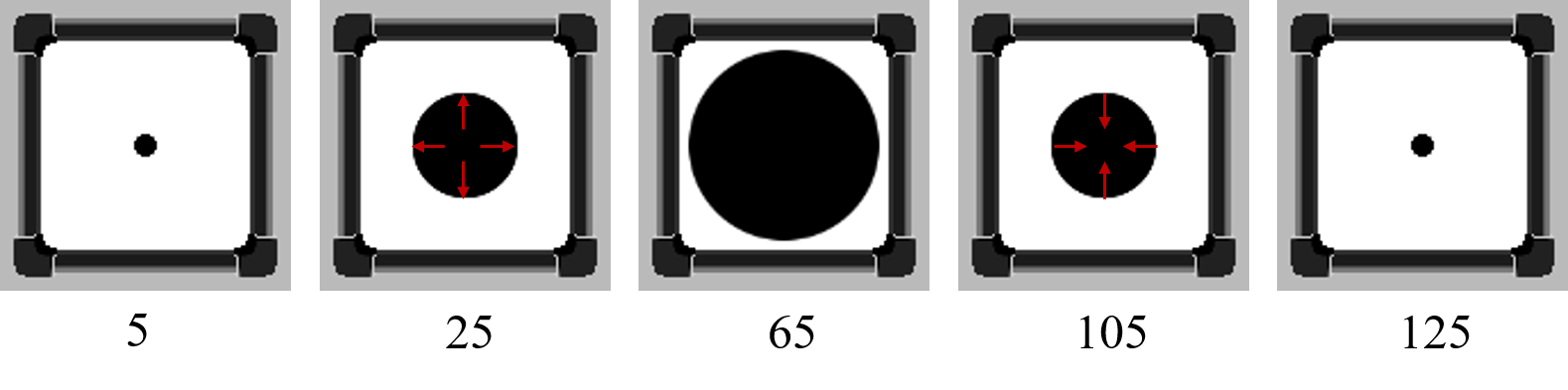}\vspace{1pt}
		\includegraphics[width=1\linewidth]{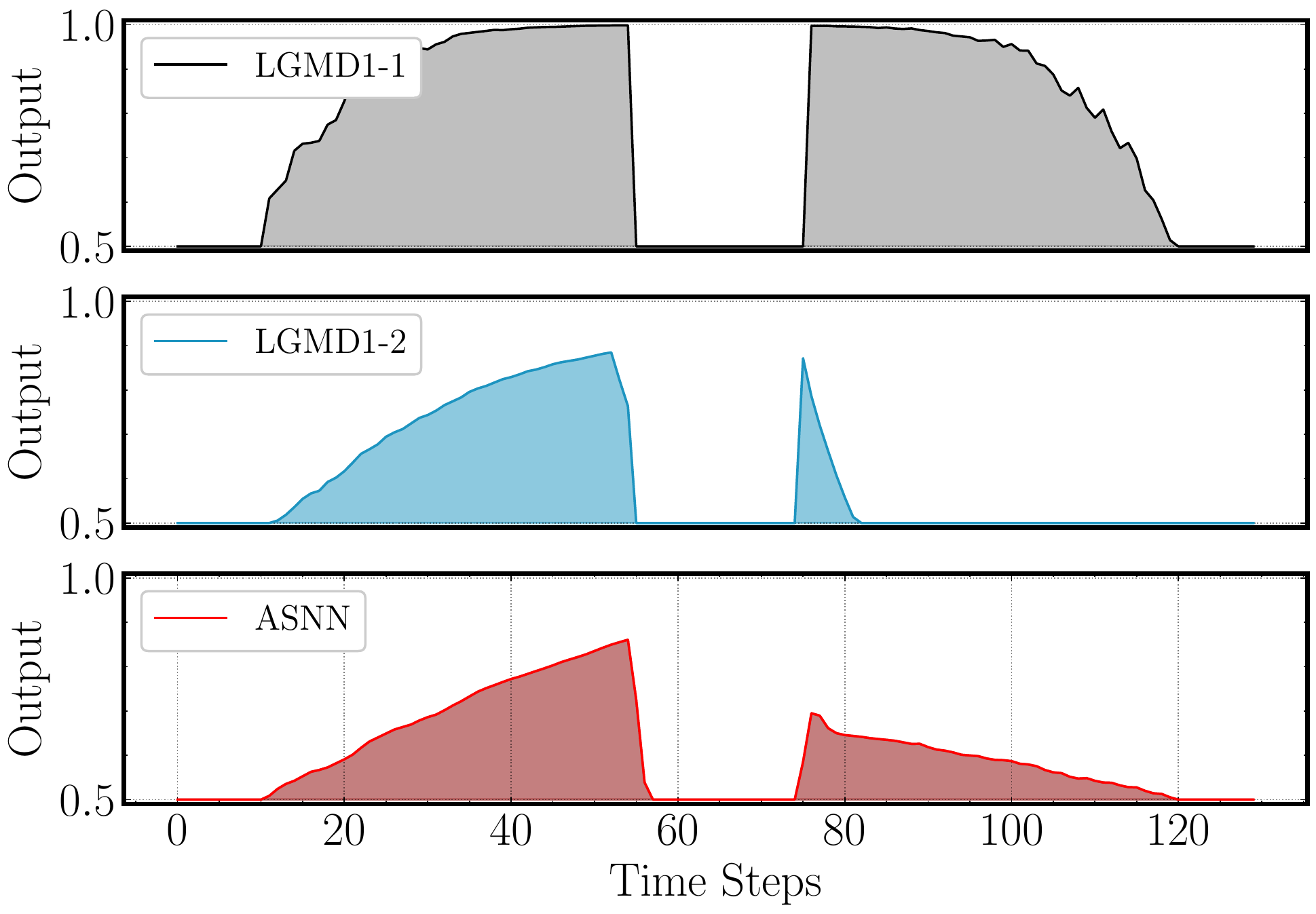}
		\end{minipage}}
	\subfigure[]{
    	\label{fig:s04f01b}
		\begin{minipage}[b]{0.3\linewidth}
		\includegraphics[width=1\linewidth]{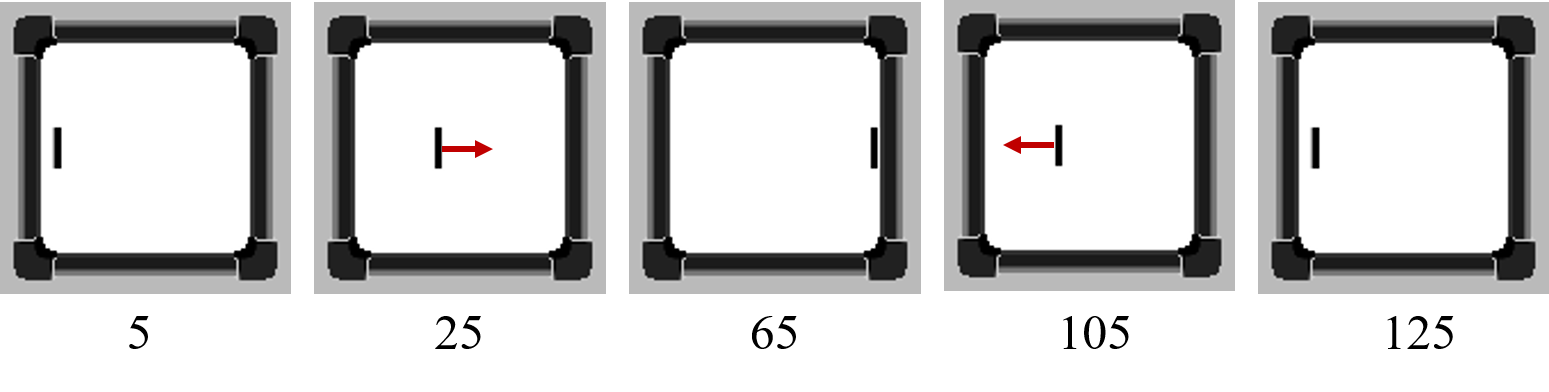}\vspace{1pt}
		\includegraphics[width=1\linewidth]{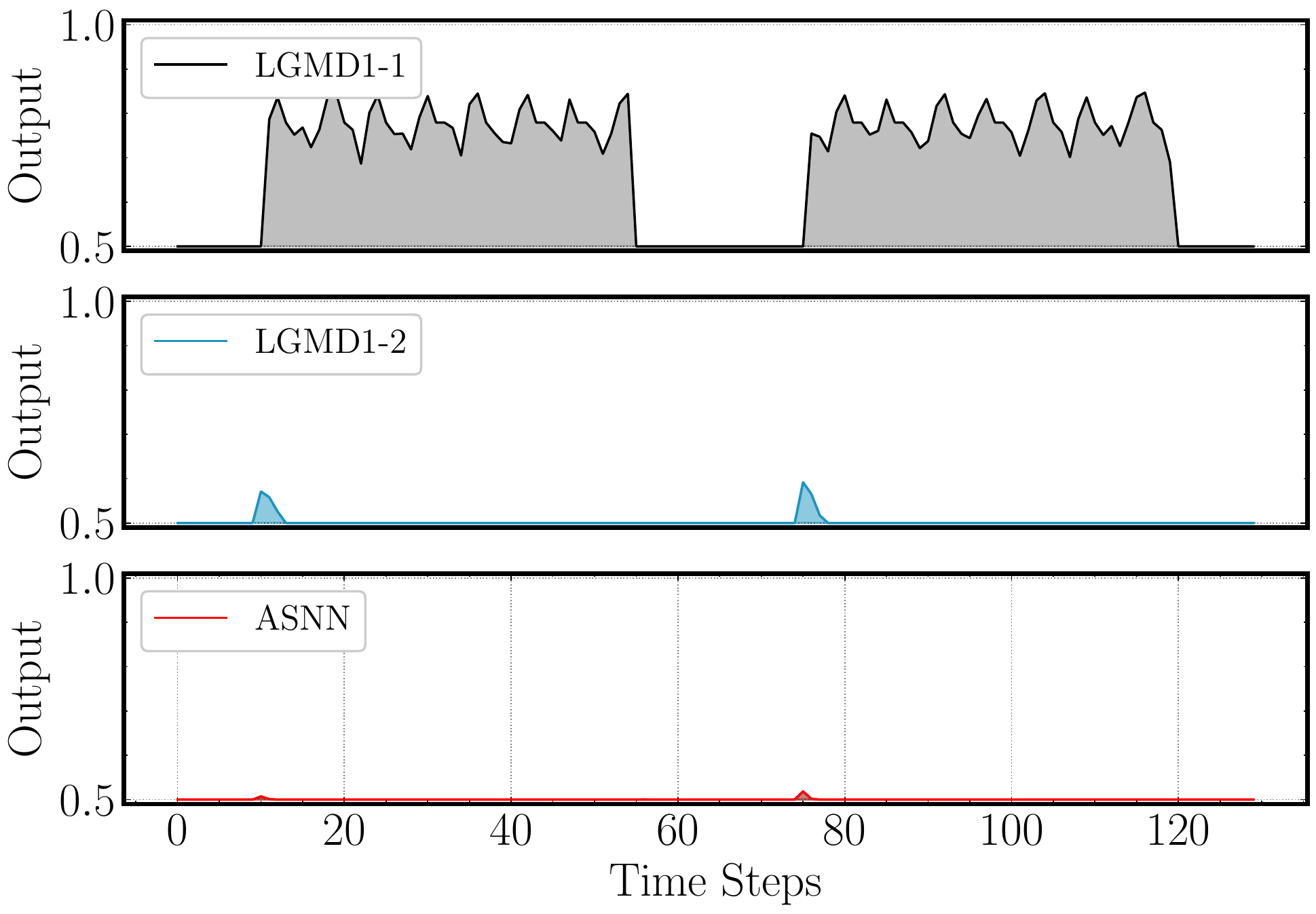}
		\end{minipage}}
	\subfigure[]{
    	\label{fig:s04f01c}
		\begin{minipage}[b]{0.3\linewidth}
		\includegraphics[width=1\linewidth]{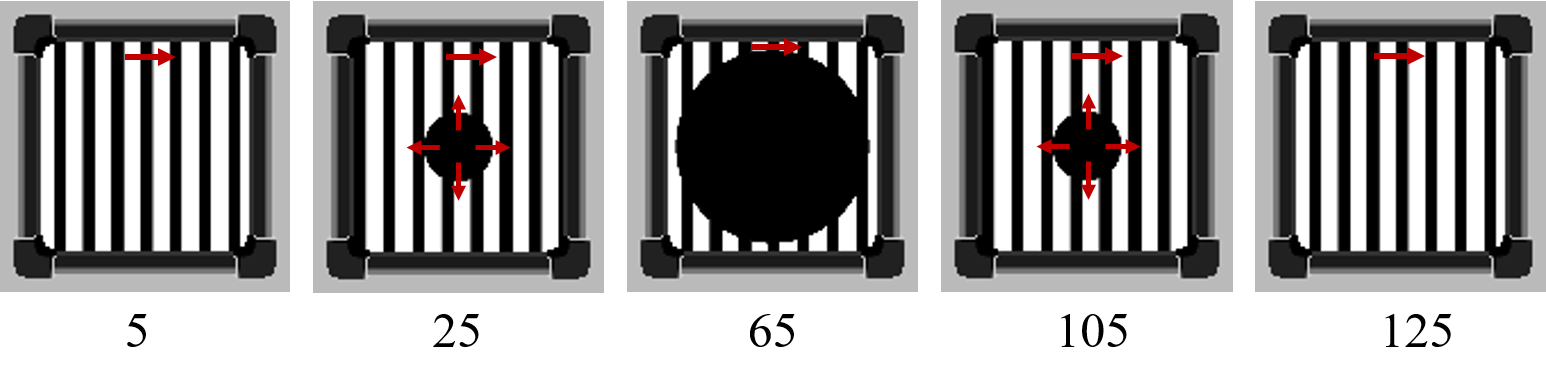}\vspace{1pt}
		\includegraphics[width=1\linewidth]{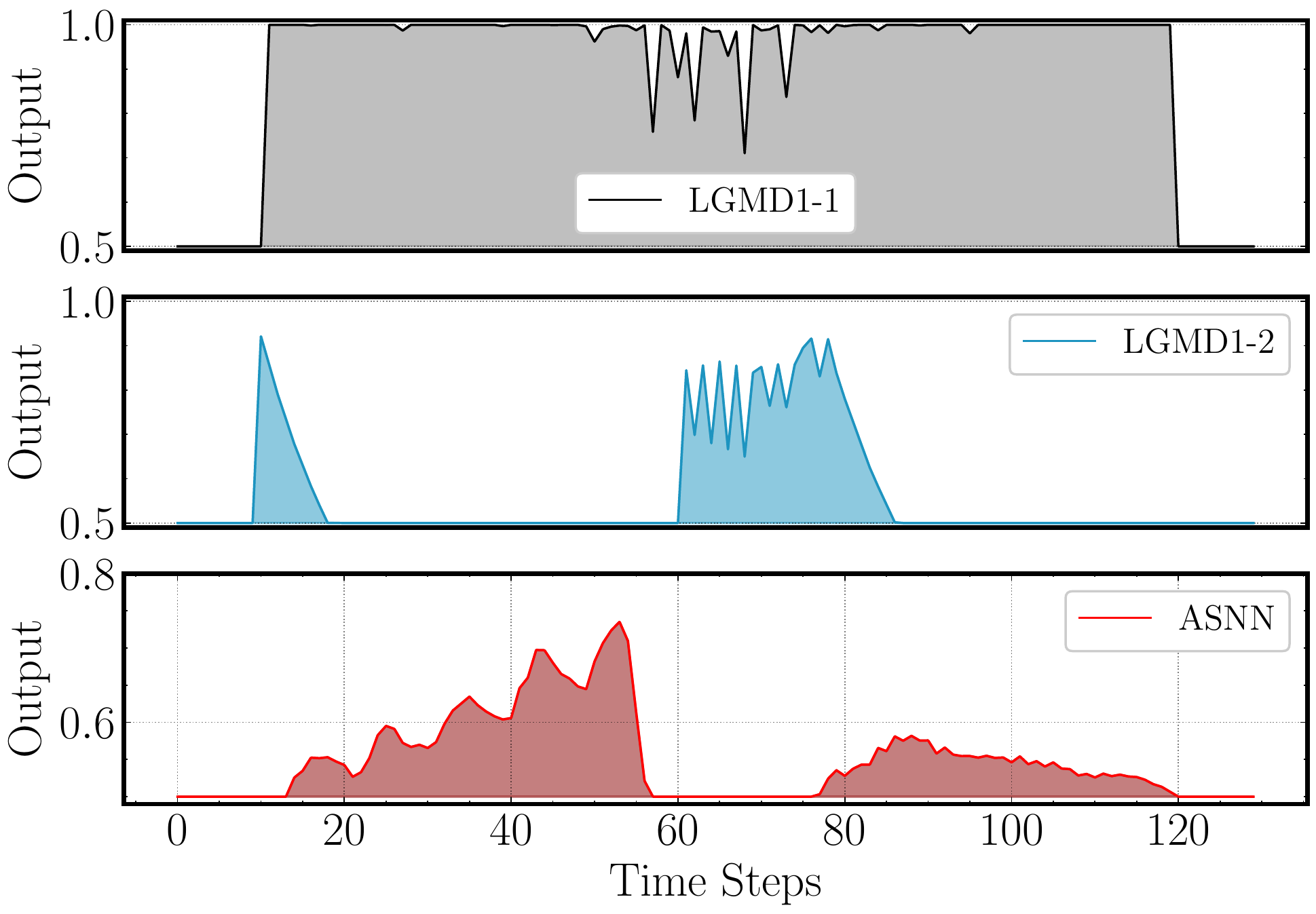}
		\end{minipage}}
    \caption{Neural responses of the proposed ASNN and the comparative LGMD model with a black target. (a) The responses in the approach-recession motion case; (b) The responses in the translation motion case; (c) The responses in the approach-recession motion case with a translating background.}
    \label{fig:s04f01}
\end{figure*}

Using the open source toolkit Psychlab developed by DeepMind, we build three synthetic environments to simulate three types of motion: approach-recession (Fig.\ref{fig:s04f01a}), translation (Fig.\ref{fig:s04f01b}) and approach-recession with a translating background (Fig.\ref{fig:s04f01c}). First of all, the approach-recession experiment is aimed at examining the sensitivity of approach without  environmental noise. The translation case is then to test the inhibitory effect on lateral motion. The final case is designed to evaluate the sensitivity of approach in a translating background, which is important for robotic visual systems to detect collision in cluttered moving backgrounds. We compare the approach sensitivity of our proposed model with the previous models LGMD1-1 \cite{yue2006collision} and LGMD1-2 \cite{fu2018shaping}.

About the experimental parameters, the input is a $128\times 128$ grayscale image in each frame. The total number of simulation steps is $T=130$, and each step lasts $\Delta t=0.05s$. More parameters of model are listed in Table.\ref{tab:s04f01}. Additionally, the sizes of kernel function $g_I$, $g_{DS}$ and $g_{CS}$ are all set as $5\times 5$, and the size of $g_{\sigma}$ is $31\times 31$.
\begin{table}[!htbp]
\caption{The predefined parameters of ASNN model} 
\centering 
\begin{tabular}{c c c c c c} 
\hline\hline 
Name & Value & Name & Value & Name & Value \\ [0.5ex] 
\hline 
$F$ & 5.0 & $\sigma_1$ & 1.0 & $\sigma_2$ & 3.0 \\ 
$K$ & 5.0 & $A$ & 60 & $\tau$ & 5 \\
$n_f$ & 2 & $n_s$ & 4 & $\sigma_{DS}$ & 0.3 \\
$\lambda_{DS}$ & 4.0 & $\sigma_{CS}$ & 0.3 & $\lambda_{CS}$ & 4.0 \\
$\sigma_g$ & 8 & $\gamma_a$ & 0.005 & $\gamma_d$ & 0.005 \\ [1ex] 
\hline 
\end{tabular}
\label{tab:s04f01} 
\end{table}

In the first synthetic test as shown in Fig.\ref{fig:s04f01}(a), a dark object expands from 10 to 55 steps and shrinks from 75 to 120 steps, which simulates the looming and receding motion respectively. Herein, the radius expanding rate of the circle is set to $s=ke^{0.001t}$ pixels/step, where $k=1.0$ is the initial rate and $t$ represents the current step. In addition, we make the weight coefficients of OFF channel larger than the ones of ON channel since most animals are more sensitive to dark looming objects. The weight coefficients are set as  $[v_{+}, v_{-}, w^1_{+},w^2_{-},w^1_{-},w^2_{+}]=[1.0, 1.0, 1.0,0.0,0.5,0.0]$. For comparing with the previous models, the neural response is also mapped into $[0.5,1.0]$ via a sigmoid activation function. Formally, the transformation is defined as follows.
\begin{equation}
out(t) = (1+\exp(\frac{|u(t)|}{N}))^{-1},
\end{equation}
where $N$ is the number of neurons. Our results in Fig.\ref{fig:s04f01}(a) show that all three methods are sensitive to dark looming objects but still with some differences. The neural response of LGMD1-1 model is symmetric in both courses of approach and recession, which is because the visual signal is processed without parallel ON and OFF pathways. When the dark object is looming, the neurons in OFF channel are activated mainly. Whereas the ON-type neurons mainly respond to the receding object. Both LGMD1-2 and our proposed model make a distinction between these two channels well. LGMD1-2 introduces a mechanism called spike frequency adaptation that enhances the LGMD's collision selectivity to looming objects rather than receding and translating stimuli. It essentially allows a neural response with a positive derivative profile to overcome adaptation selectively and blocks other responses. However, that method is only directed to the output node that represents the sum of all neural responses, which is difficult to extract more detailed information. For example, if multiple targets appear at the same time, some approaching and some receding, it becomes difficult to determine whether a collision will happen and where the looming objects are. Our proposed framework can control the response intensity of different channels by adjusting the weight coefficients. Meanwhile, lateral motion inhibition and directionally selective inhibition are executed in a pixel-wise manner. 
\begin{figure*}[!htbp]
    \centering
    \subfigure[]{
    	\label{fig:s04f02a}
		\begin{minipage}[b]{0.3\linewidth}
		\includegraphics[width=1\linewidth]{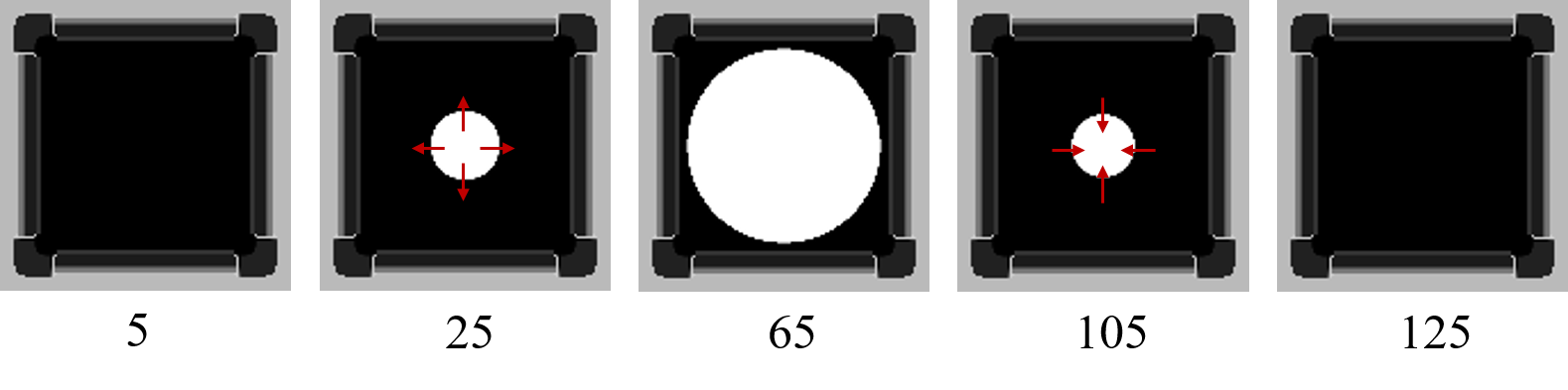}\vspace{1pt}
		\includegraphics[width=1\linewidth]{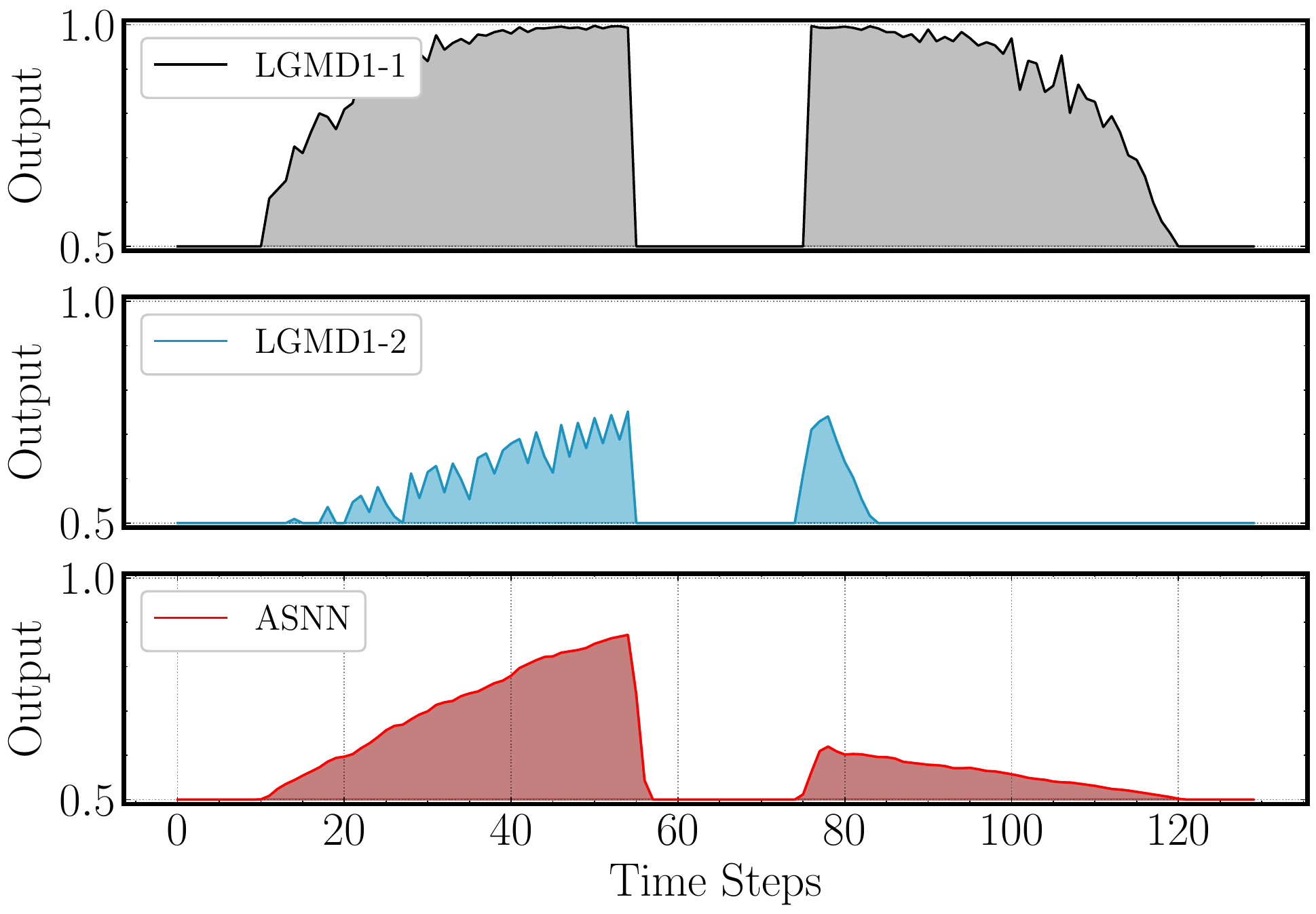}
		\end{minipage}}
	\subfigure[]{
    	\label{fig:s04f02b}
		\begin{minipage}[b]{0.3\linewidth}
		\includegraphics[width=1\linewidth]{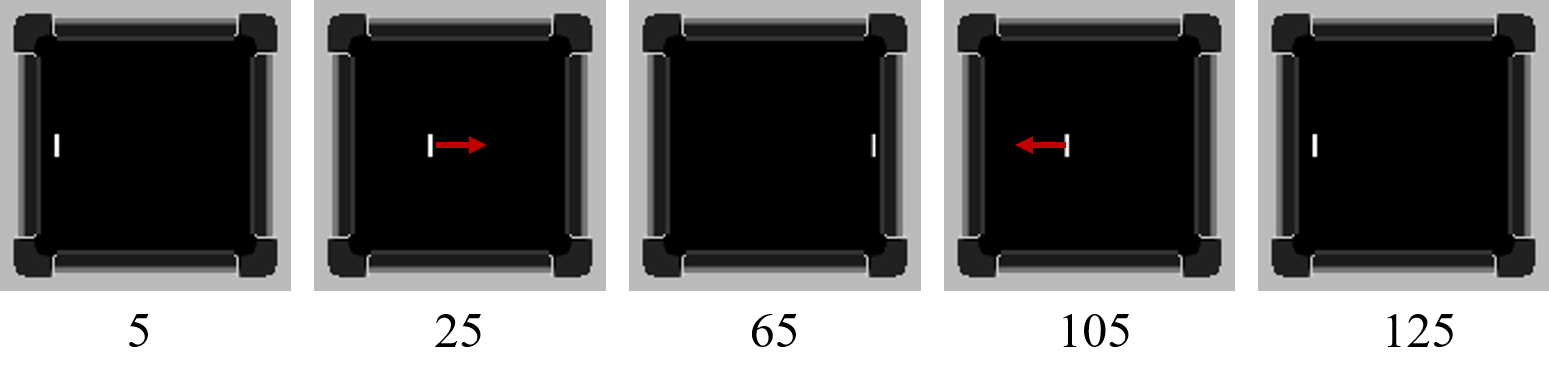}\vspace{1pt}
		\includegraphics[width=1\linewidth]{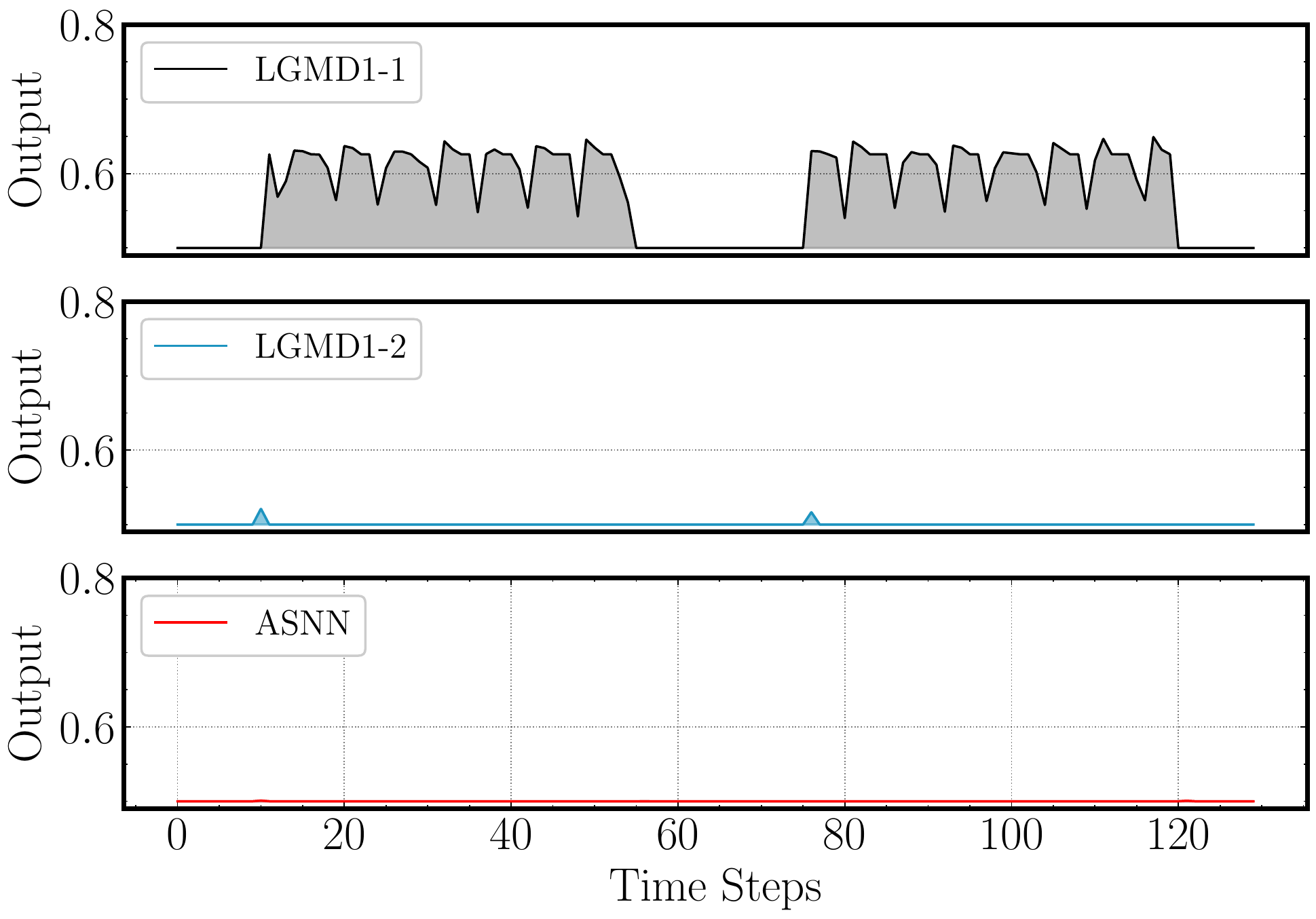}
		\end{minipage}}
	\subfigure[]{
    	\label{fig:s04f02c}
		\begin{minipage}[b]{0.3\linewidth}
		\includegraphics[width=1\linewidth]{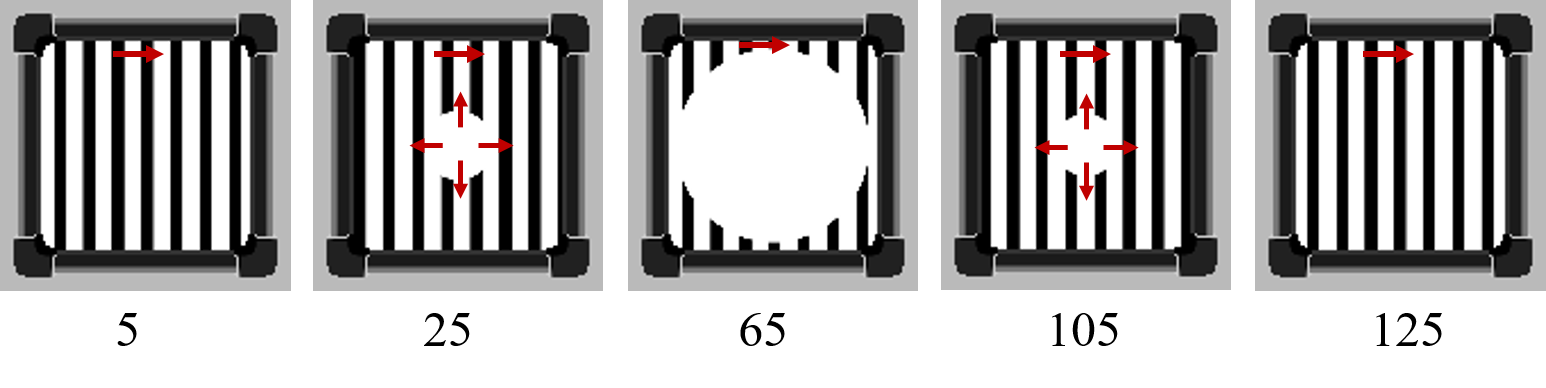}\vspace{1pt}
		\includegraphics[width=1\linewidth]{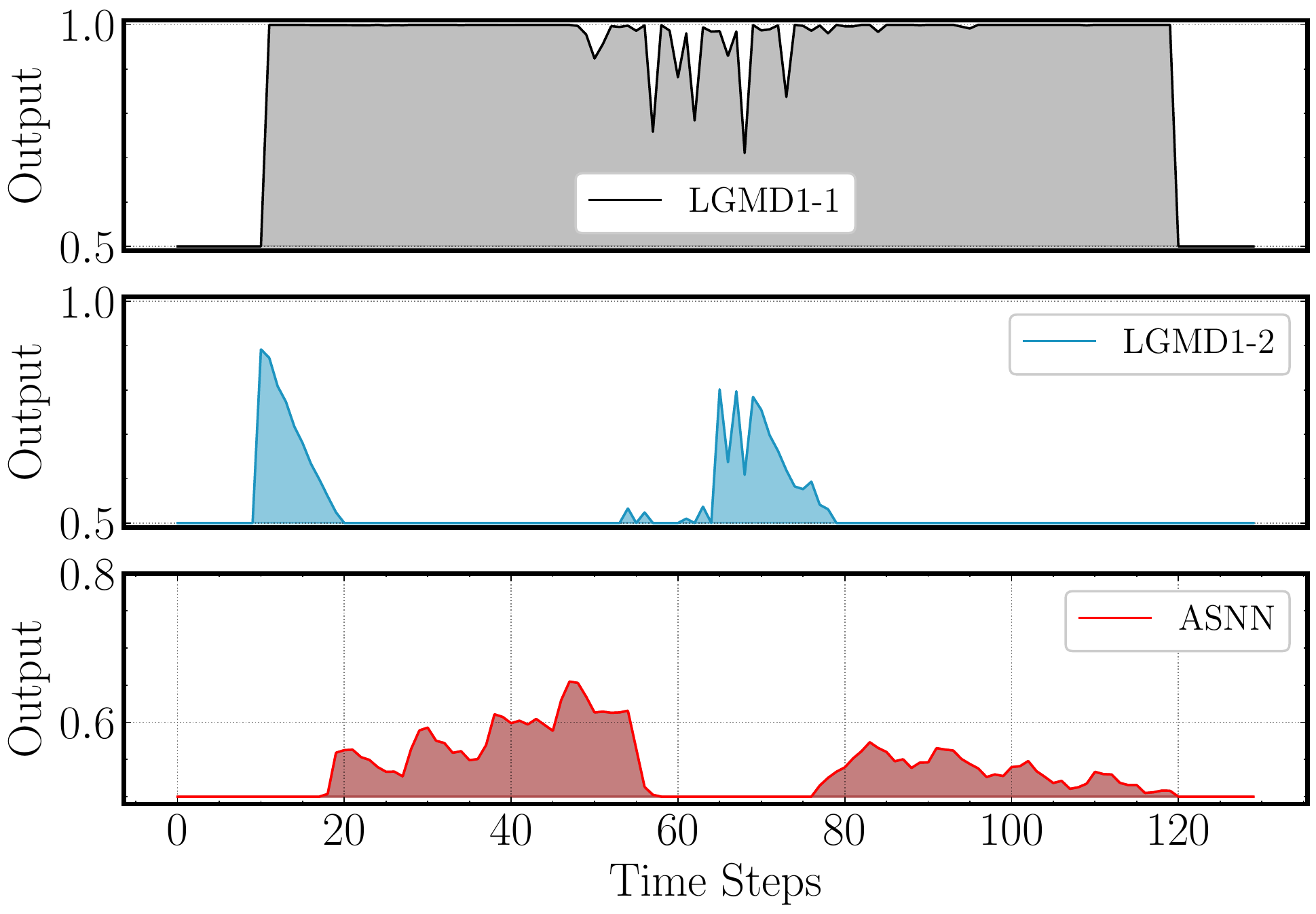}
		\end{minipage}}
    \caption{Neural responses of the proposed ASNN and the comparative LGMD model with a white target. (a) The responses in the approach-recession motion case; (b) The responses in the translation motion case; (c) The responses in the approach-recession motion case with a translating background.}
    \label{fig:s04f02}
\end{figure*}

In the second synthetic test (Fig.\ref{fig:s04f01}(b)), we investigate the neural responses of three models in the case of translating movements. A dark bar moves right from 10 to 55 steps and moves left from 75 to 120 steps, with a constant speed $2$ pixels/step. The weight coefficients are set as $[v_{+}, v_{-}, w^1_{+},w^2_{-},w^1_{-},w^2_{+}]=[1.0, 1.0, 0.4,1.0,0.0,0.0]$. Obviously, the LGMD1-1 model produces a larger neural responses without any decay, which demonstrates that it fails to inhibit the lateral motion or can not distinguish the radial and lateral motion well. The LGMD1-2 has a weak and quickly decayed response to translations at constant speeds. The proposed model demonstrates a much weaker response than other two comparative models. As mentioned above, this is because the slow 'leading' responses from OFF-Bipolar cells are mostly inhibited by the fast 'trailing' responses from ON-SACs, and the fast 'trailing' responses from ON-Bipolar cells are mostly inhibited by the slow 'leading' responses from OFF-SACs. But this lateral motion inhibition is limited since the receptive field of antagonistic center-surround spatial filtering is relatively small. If the object is very wide along motion direction, the inhibitory area produced by the SACs can not completely cover the excitatory area of the Bipolar cells. In that case, we should turn to directionally selective inhibition. 

In the actual implementation, the translating objects have a serious impact on collision detection, which may cause many false positives. Thus, in the third synthetic test (Fig.\ref{fig:s04f01}(c)), the approach sensitivity in a moving background is investigated further. A lot of black vertical stripes fill the background, and the foreground is a looming dark disk same as the first experiment. During the simulation, all stripes move from left to right. The experimental results demonstrate that all previous models fail to tackle this challenging problem. The LGMD1-1 almost maintains a stronger response all the time since it cannot distinguish between lateral and radial motion. Likewise, the LGMD1-2 cannot overcome the interference of moving background as well. The change in the background does not always keep the sum of neural responses in a positive derivative profile when the object approaching, which makes the spike frequency adaptation algorithm not work. While our proposed model is robust to detect the approaching process in the moving background. The neural response is approximately consistent with the first experiment, which mainly benefits from the process of directionally selective inhibition. In addition, we further reverse the settings of the target and background to evaluate the performance of motion perception. As shown in Fig.\ref{fig:s04f02}, the results are similar. 

\begin{figure*}[!htbp]
    \centering
    \includegraphics[width=0.95\textwidth]{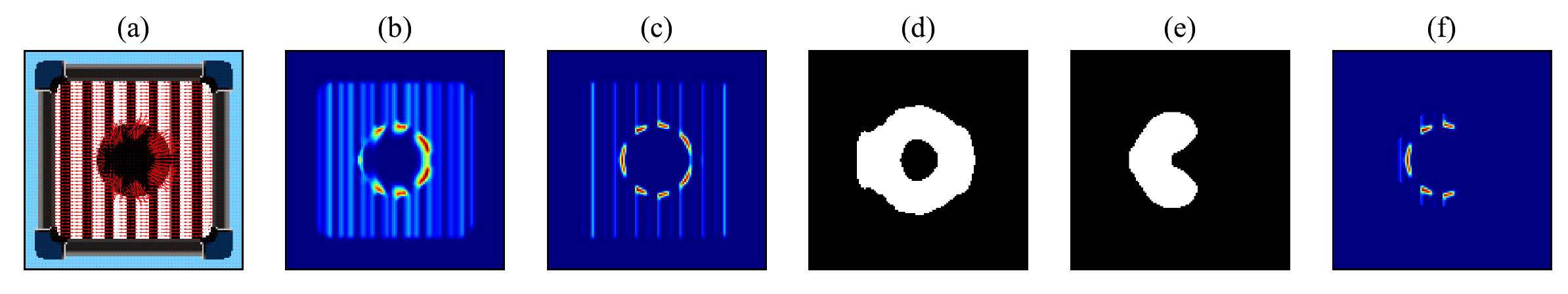}
    \caption{(a) The flow of motion energy. (b) Maximal motion energy $V$. (c) Approach-sensitive Map $G$. (d) Approach-sensitive Attention $M_a$. (e) Directionally Selective Attention Mask $M_d$. (f) Approach-sensitive Response $G'$}
    \label{fig:s04f04}
\end{figure*}

For the approach-recession motion against a translating background, neural responses of the intermediate layers at $32th$ step are depicted detailedly in Fig.\ref{fig:s04f04}. Firstly, the direction perception algorithm is able to estimate the motion directions of targets and backgrounds accurately (Fig.\ref{fig:s04f04}(a)). All stripes are detected as moving from left to right, and the disk is expanding along all directions. Fig.\ref{fig:s04f04}(b) depicts the maximal motion energy $V$ that gathers the filtering results in all directions. Fig.\ref{fig:s04f04}(c) shows the approach-sensitive responses $G$ of retinal ganglion cells, in which the lateral motion is weaken clearly due to the antagonistic center-surround filtering. But the wide-field lateral motion is still not suppressed effectively such that the approach-sensitive attention $M_1$ is affected by part of the moving background (Fig.\ref{fig:s04f04}(d)). While the directionally selective inhibitation plays a key role in suppressing the moving background since all targets moving from left to right are removed (Fig.\ref{fig:s04f04}(e)). For the expanding disk, the neural responses are inhibited in only one direction as shown in Fig.\ref{fig:s04f04}(f), which has little impact on detecting collision. 

\begin{figure}[!htbp]
    \centering
    \subfigure[]{\label{fig:s04f05a}\includegraphics[width=0.45\textwidth]{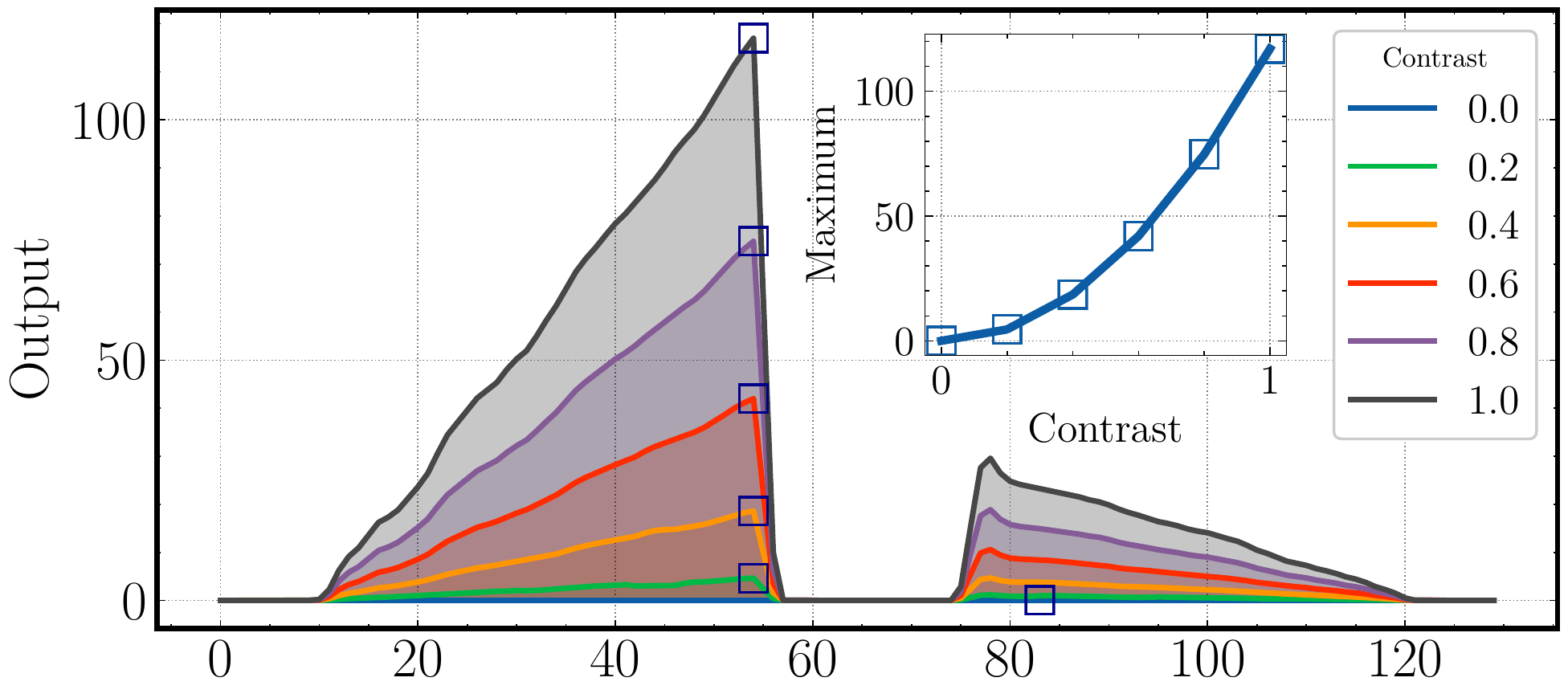}}\\
    \subfigure[]{\label{fig:s04f05b}\includegraphics[width=0.45\textwidth]{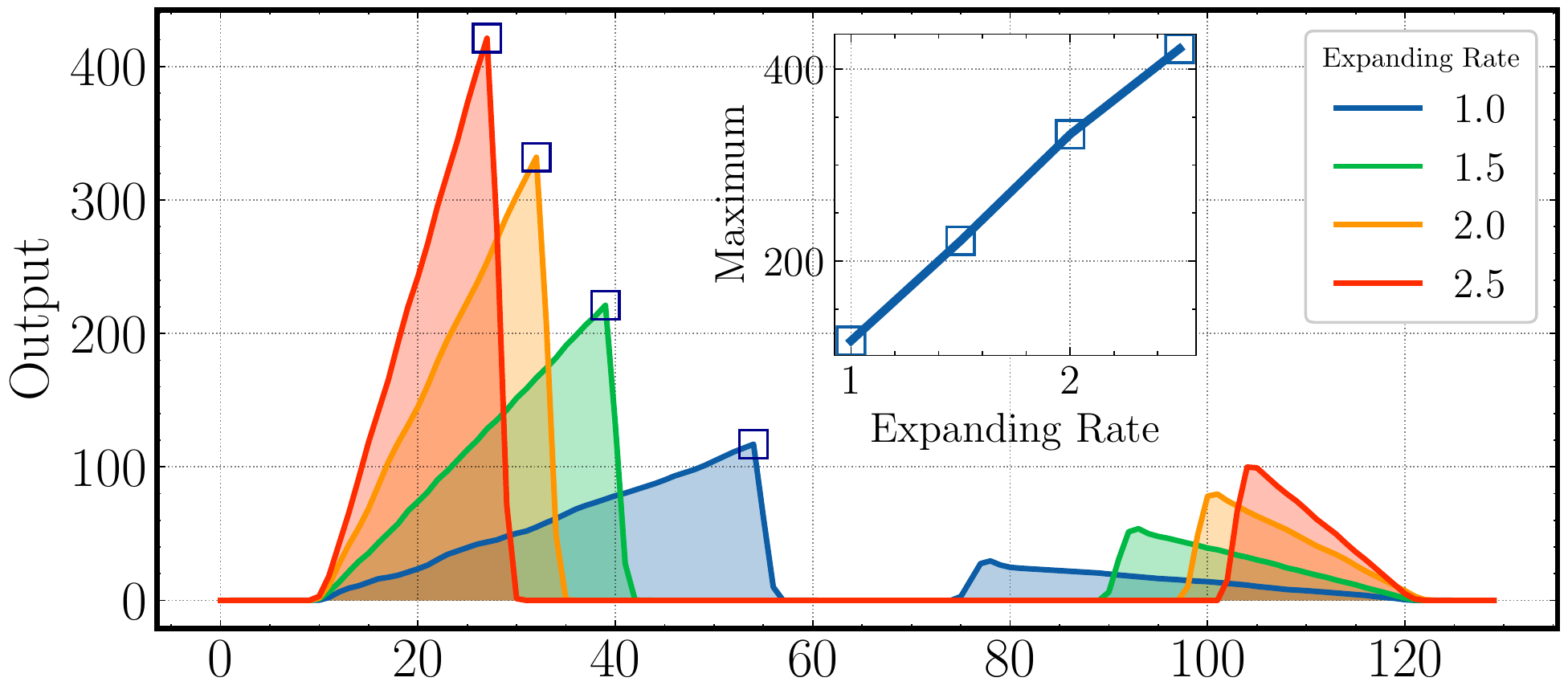}}
    \caption{The effects of contrast and speed on approach sensitivity. (a) Neural responses of the proposed ASNN at different Weber contrast. (b) Neural responses of the proposed ASNN at different expanding rate.}
    \label{fig:s04f05}
\end{figure}

Furthermore, we investigate the effects of contrast and speed on approach sensitivity based on the first synthetic test. Firstly, Weber contrast is used to examine the contrast sensitivity, defined as $c=\frac{|\bar{I}_t-\bar{I}_b|}{255}$. $\bar{I}_t$ is the average gray-scale value of targets, and $\bar{I}_b$ is the average gray-scale value in the neighboring area around the target. Fixing the edge expanding rate to 1.0 pixel/step, the outputs $u(t)$ and their peaks at different Weber contrast are recorded in Fig.\ref{fig:s04f05}(a). As the contrast increases, the responsive intensity grows approximately exponentially. A high-contrast looming object is easier to be detected. After that, different expanding rates are chosen to examine the speed sensitivity. The Weber contrast is fixed to 1.0. The neural responses and peaks are summarized in Fig.\ref{fig:s04f05}(b). It is clear that the increasing expanding rate can enhance the strength of the response linearly. The faster the object approaches, the easier it is to be detected.

\subsection{Approach Sensitivity in Real Scenarios}
As we know, approach sensitivity is significent for vision-based collision avoidance systems in the ground vehicle applications. The proposed model is further tested in real on-road videos from vehicle camera recorders, for assessing the performance of collision detection. Herein, the dataset is from Fu's work \cite{fu2020improved}, which includes some colliding and non-colliding driving scenarios on crowded urban road during the day or night. 
Similar to the previous models, we also compute the spike frequency as the indicator of potential collision threat. More specifically, the number of spikes firstly could be generated by an exponential mapping from membrane potential to the firing rate as follows.
\begin{equation}
spike(t) = \lfloor e^{[K_{sp}(out(t)-T_{sp})]} \rfloor,
\end{equation}
where $\lfloor \cdot \rfloor$ is a floor function. $K_{sp}$ and $T_{sp}$ represent a scale parameter and a threshold parameter respectively. Then the collision warning is generated by  comparing the spike frequency with a given threshold.
\begin{align}
\begin{split}
Col(t)= \left \{
\begin{array}{ll}
True, & \frac{1}{N_t\Delta t} \sum_{i-N_t}^t spike(t) \geq T_c \\
False, & otherwise
\end{array}
\right. ,
\end{split}
\end{align}
where $N_t$ and $T_c$ denotes the window size and detection threshold.

\begin{figure*}[!htbp]
    \centering
    \subfigure[]{
    	\label{fig:s04f06a}
		\begin{minipage}[b]{0.3\linewidth}
		\includegraphics[width=1\linewidth]{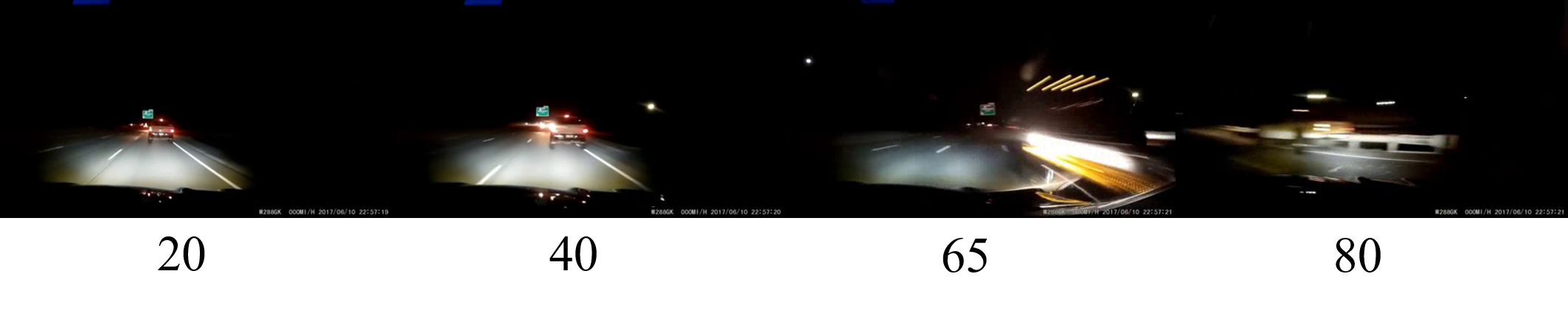}\vspace{1pt}
		\includegraphics[width=1\linewidth]{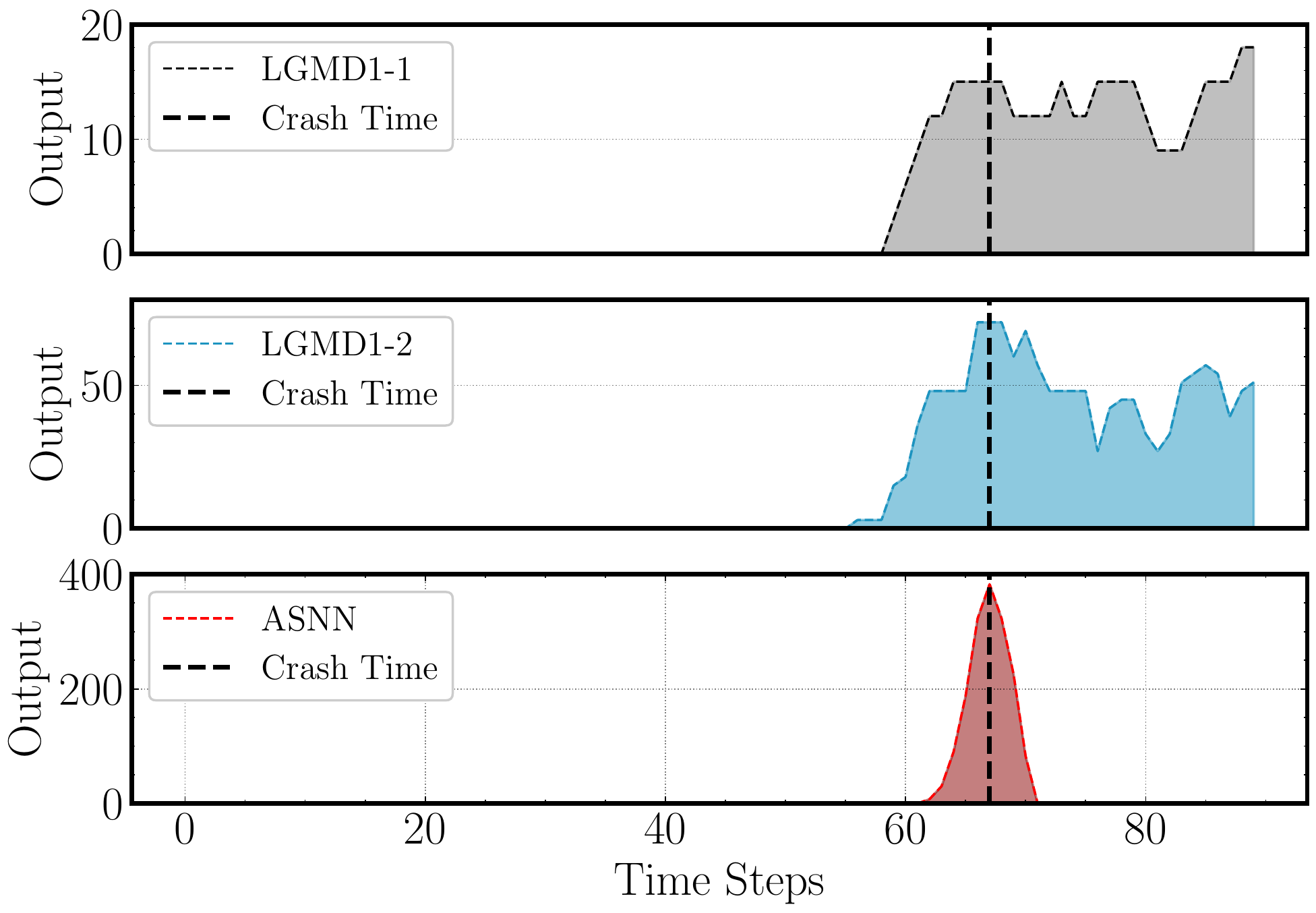}
		\end{minipage}}
	\subfigure[]{
    	\label{fig:s04f06b}
		\begin{minipage}[b]{0.3\linewidth}
		\includegraphics[width=1\linewidth]{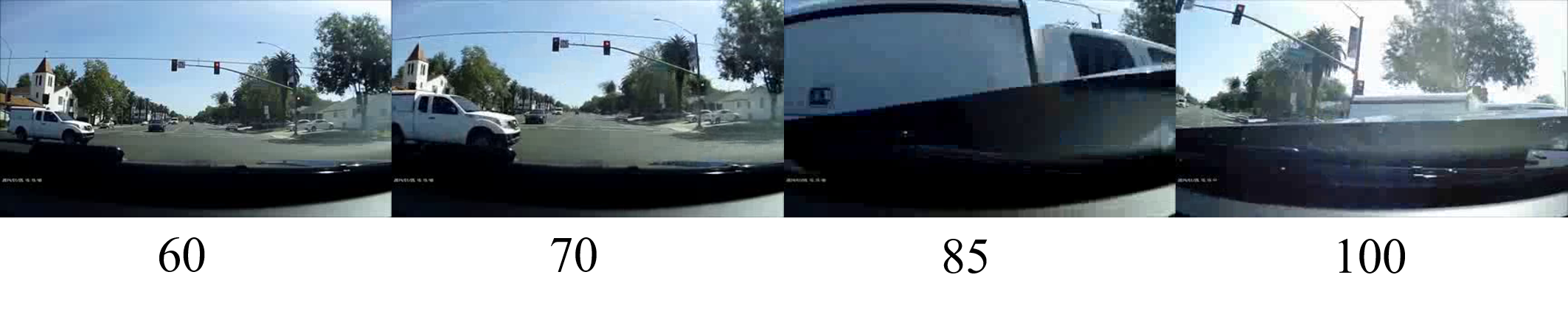}\vspace{1pt}
		\includegraphics[width=1\linewidth]{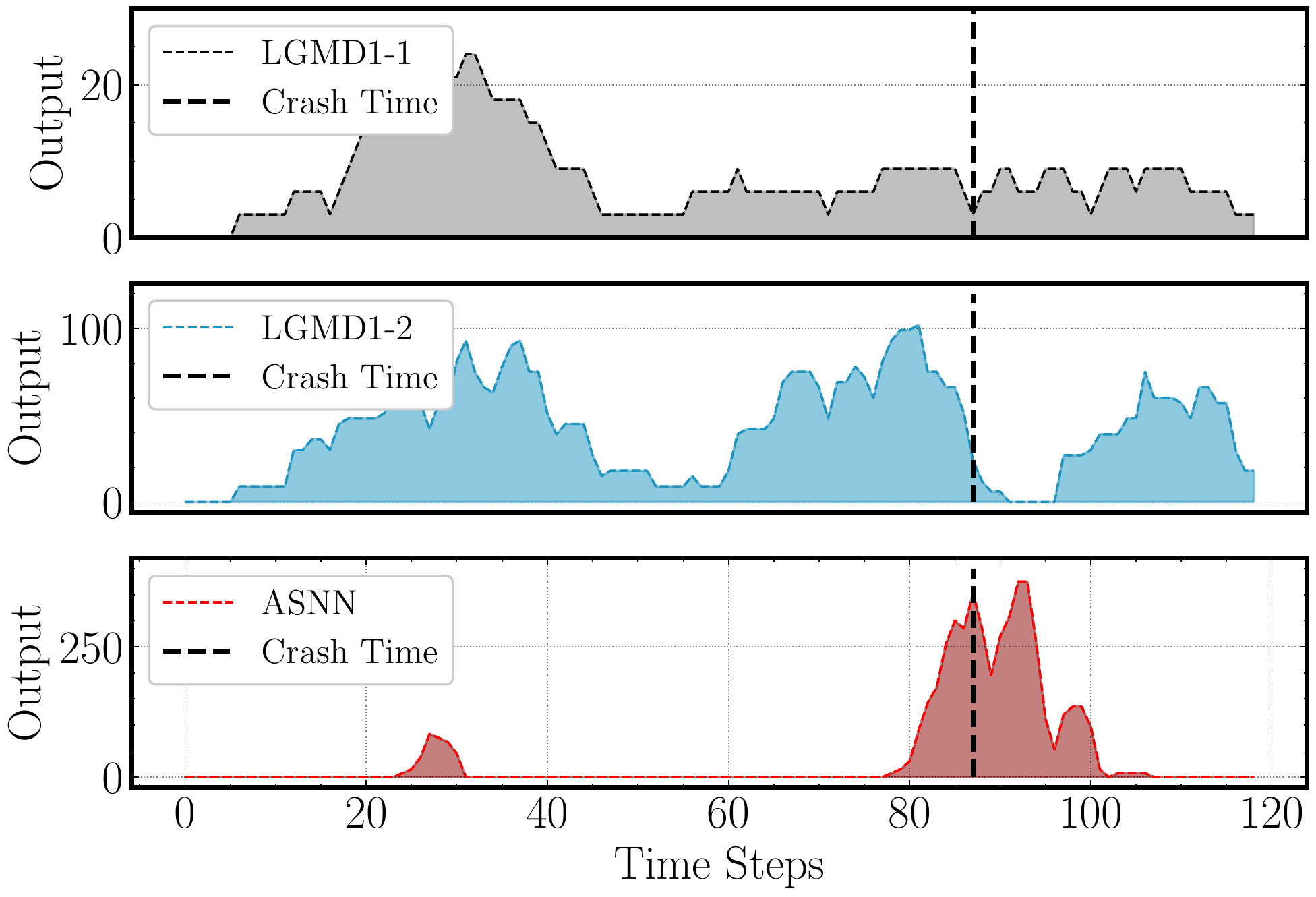}
		\end{minipage}}
	\subfigure[]{
    	\label{fig:s04f06c}
		\begin{minipage}[b]{0.3\linewidth}
		\includegraphics[width=1\linewidth]{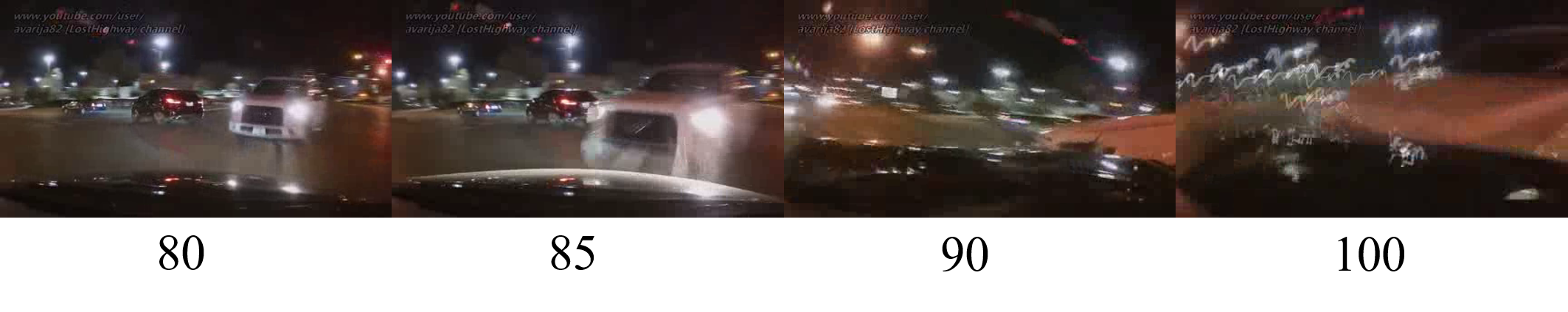}\vspace{1pt}
		\includegraphics[width=1\linewidth]{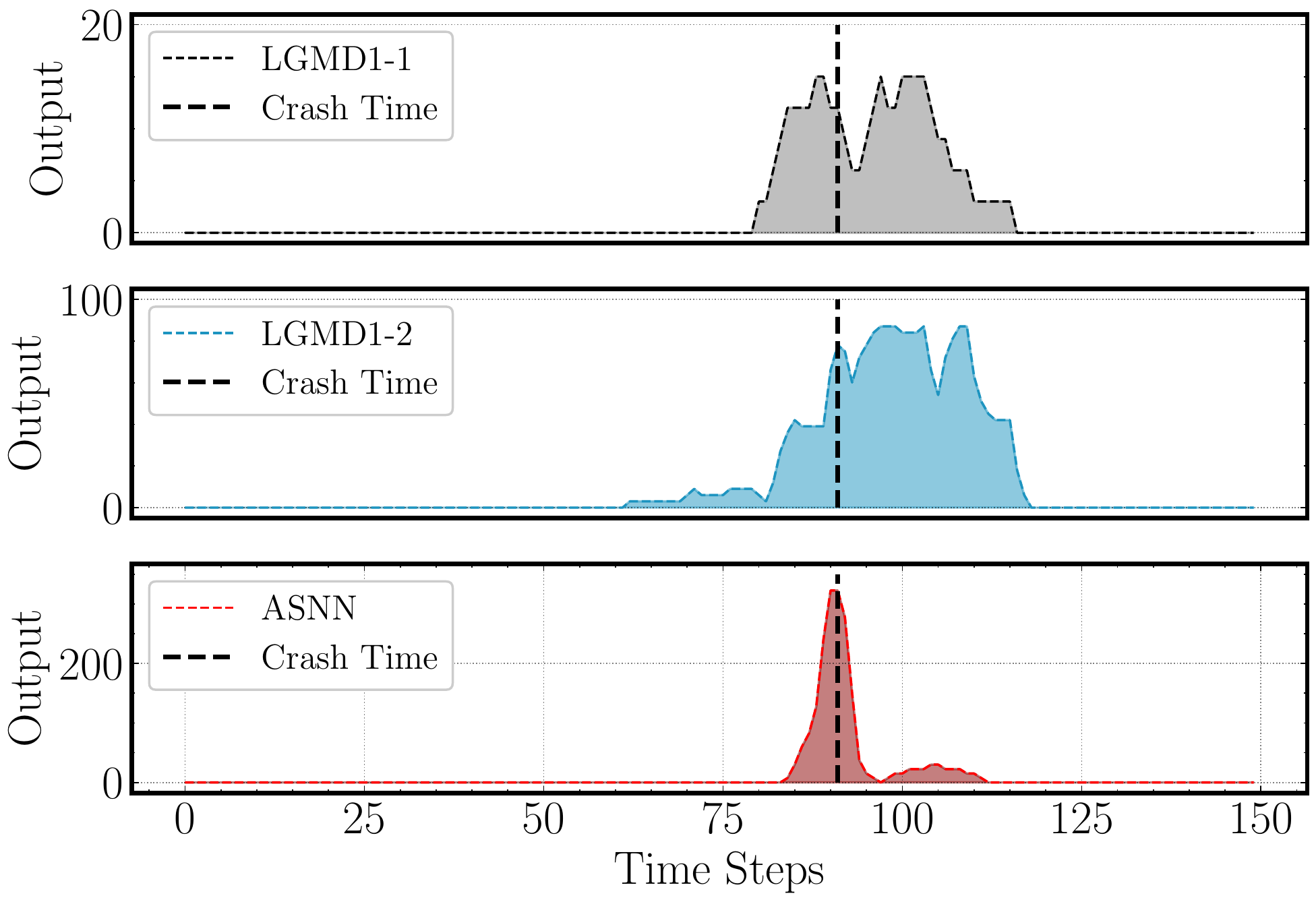}
		\end{minipage}}
	\subfigure[]{
    	\label{fig:s04f06d}
		\begin{minipage}[b]{0.3\linewidth}
		\includegraphics[width=1\linewidth]{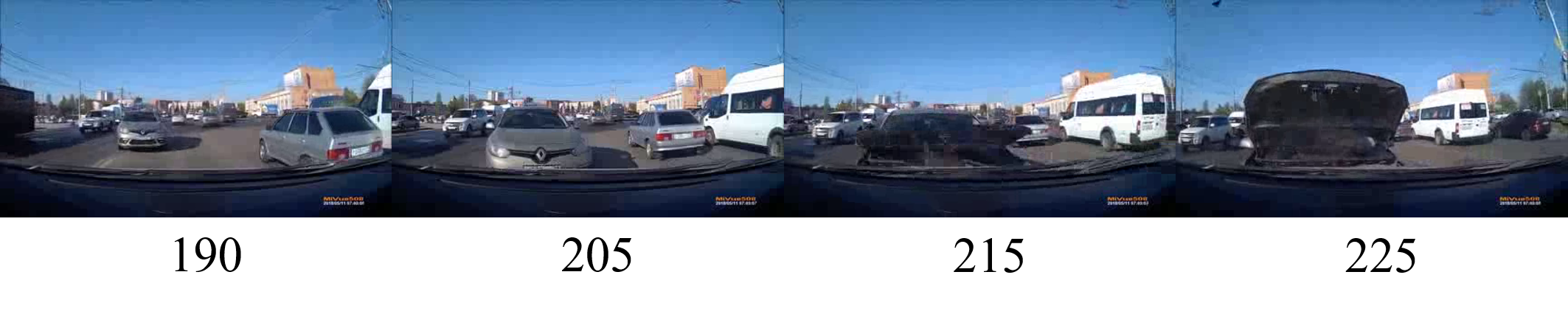}\vspace{1pt}
		\includegraphics[width=1\linewidth]{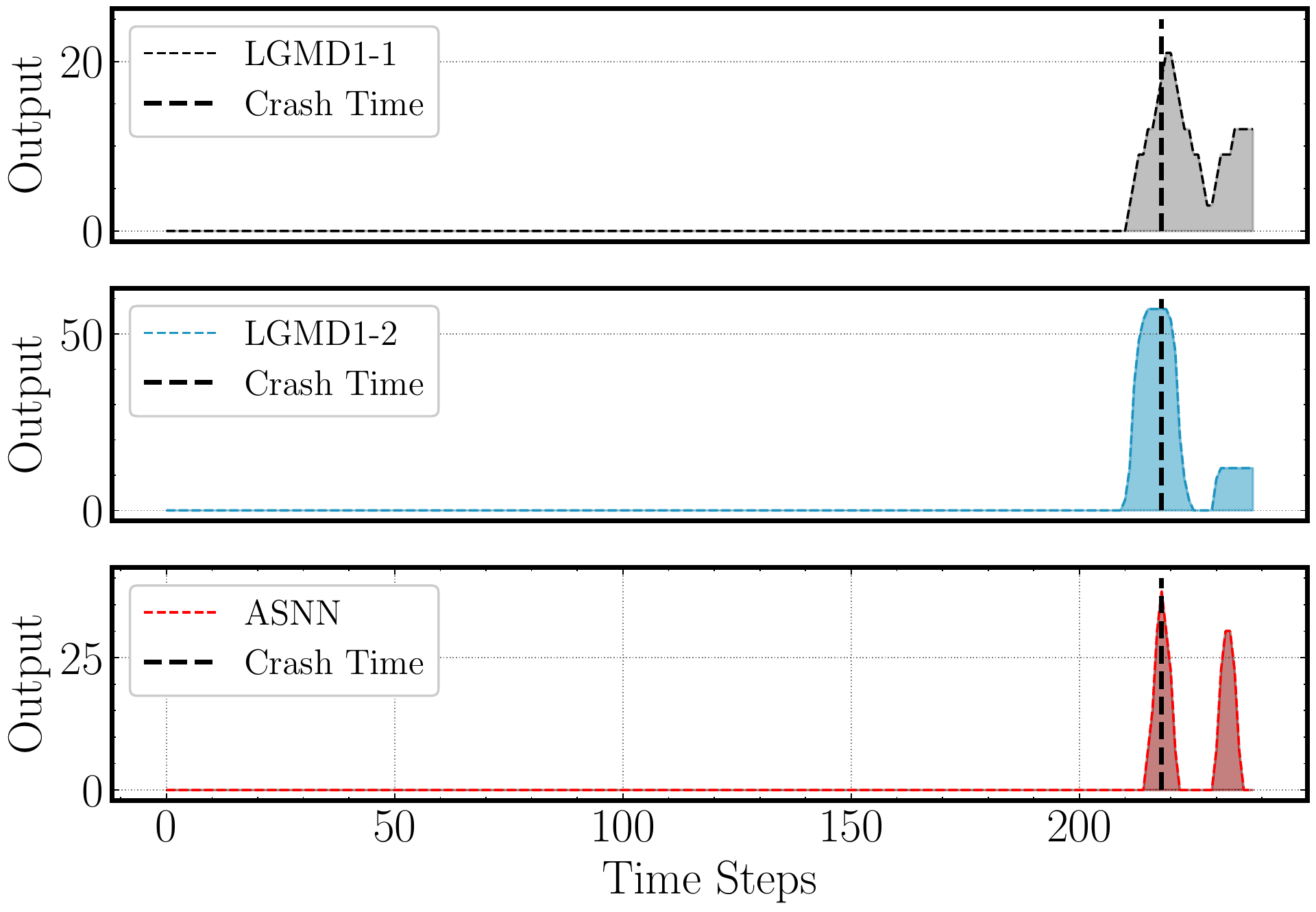}
		\end{minipage}}
	\subfigure[]{
    	\label{fig:s04f06e}
		\begin{minipage}[b]{0.3\linewidth}
		\includegraphics[width=1\linewidth]{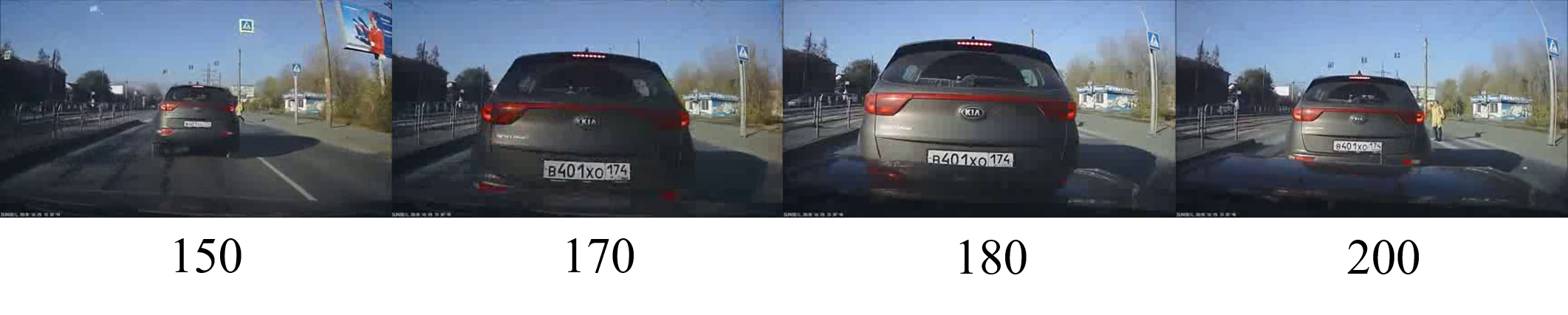}\vspace{1pt}
		\includegraphics[width=1\linewidth]{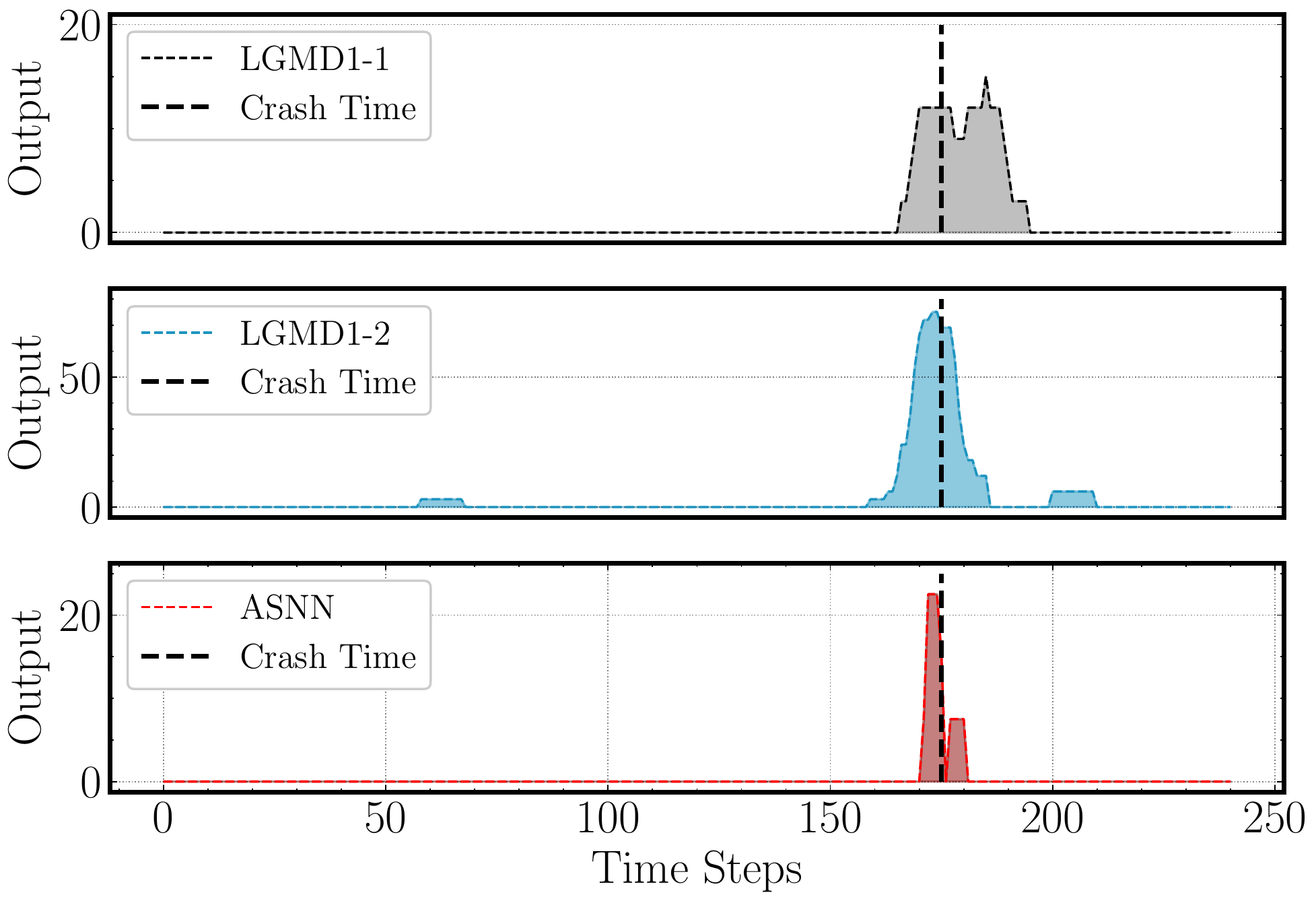}
		\end{minipage}}
	\subfigure[]{
    	\label{fig:s04f06f}
		\begin{minipage}[b]{0.3\linewidth}
		\includegraphics[width=1\linewidth]{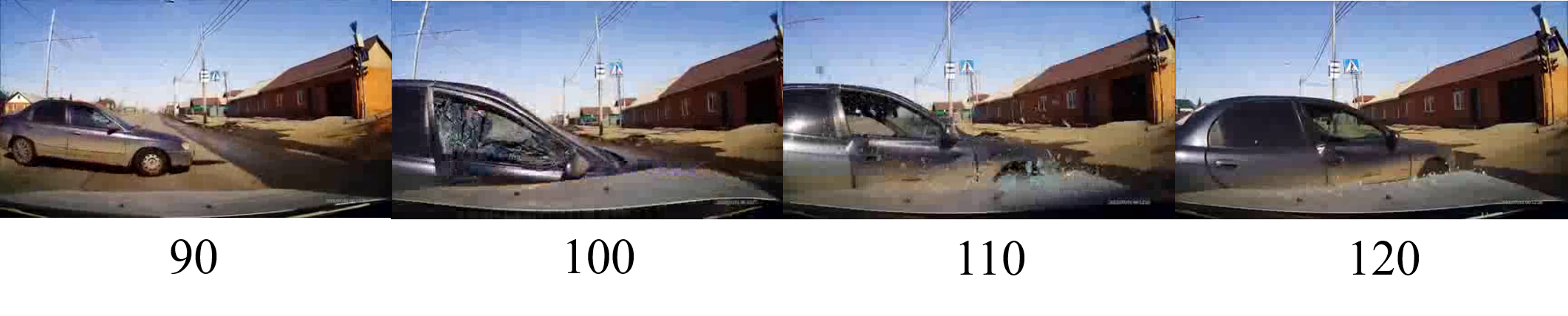}\vspace{1pt}
		\includegraphics[width=1\linewidth]{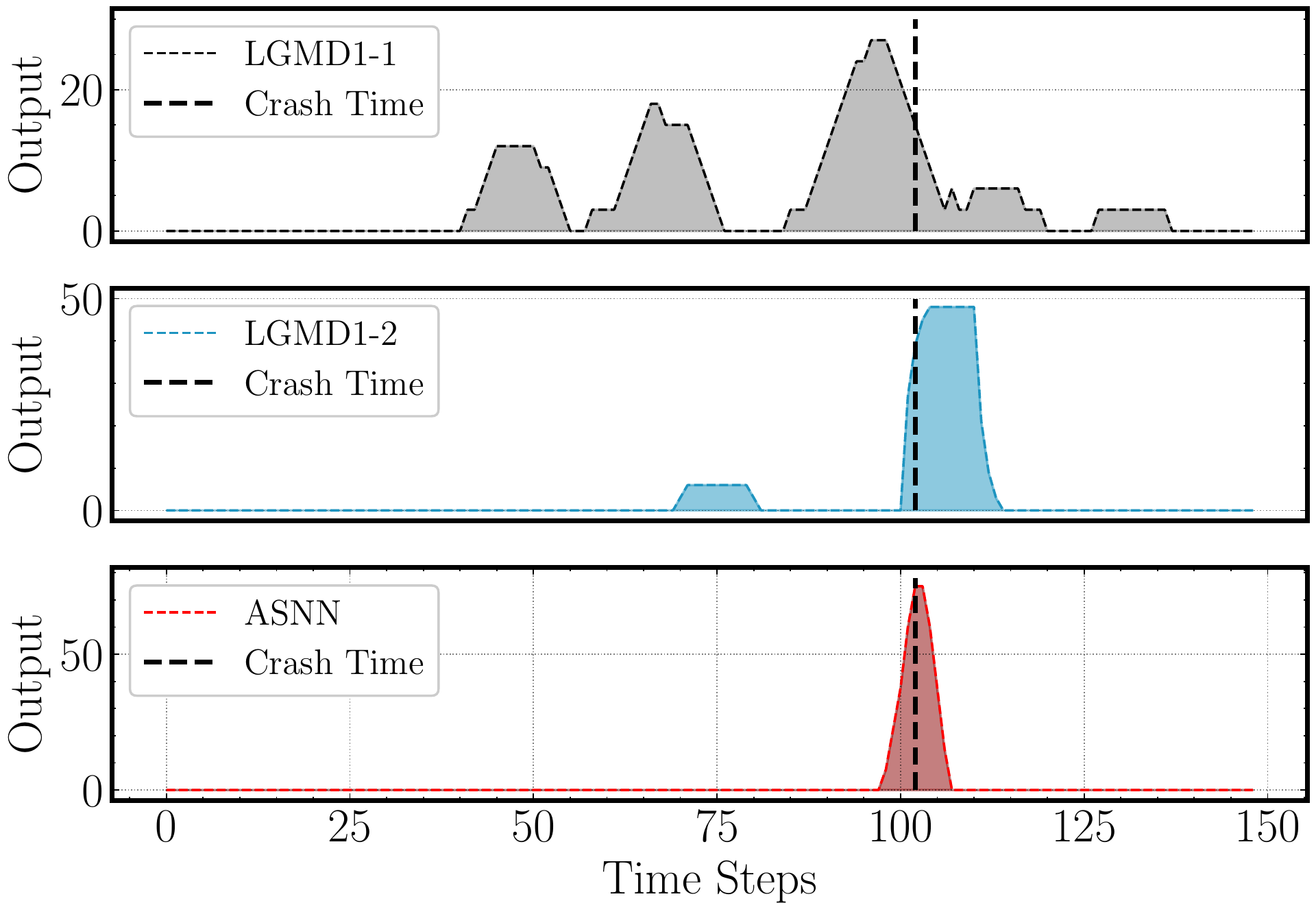}
		\end{minipage}}
    \caption{The neural responses of the proposed ASNN and the comparative LGMD model in six collision scenarios.}
    \label{fig:s04f06}
\end{figure*}

\begin{figure*}[!htbp]
    \centering
	\subfigure[]{
    	\label{fig:s04f06g}
		\begin{minipage}[b]{0.3\linewidth}
		\includegraphics[width=1\linewidth]{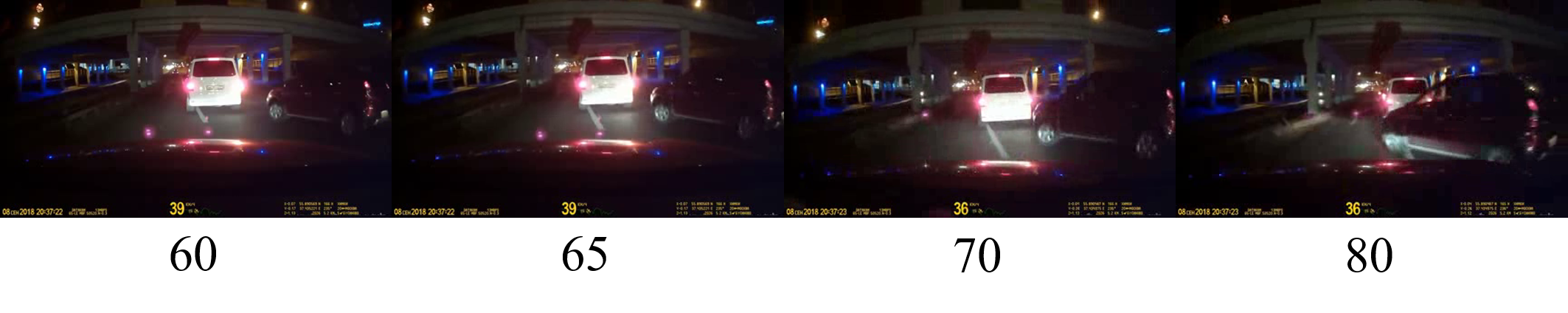}\vspace{1pt}
		\includegraphics[width=1\linewidth]{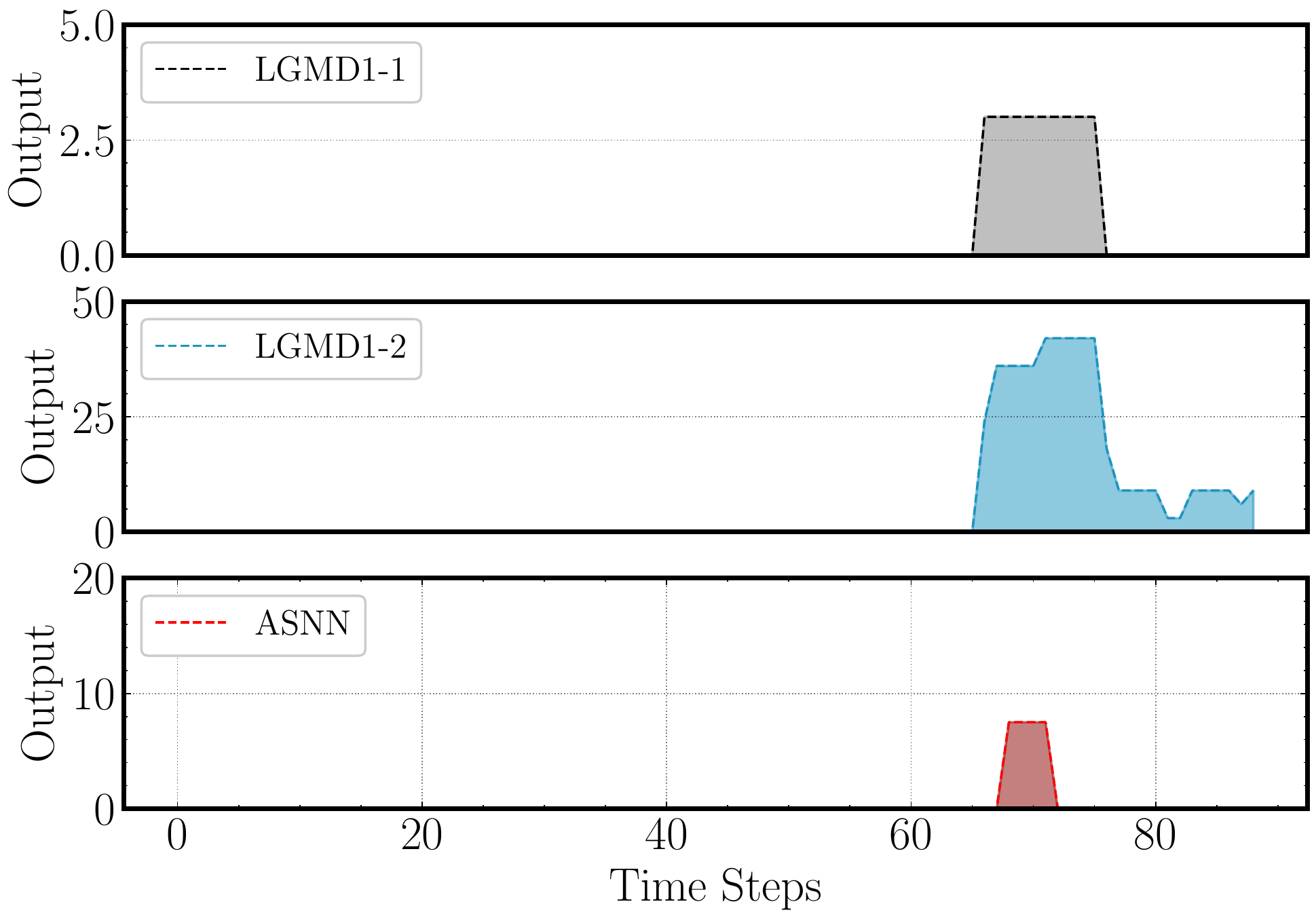}
		\end{minipage}}
	\subfigure[]{
    	\label{fig:s04f06h}
		\begin{minipage}[b]{0.3\linewidth}
		\includegraphics[width=1\linewidth]{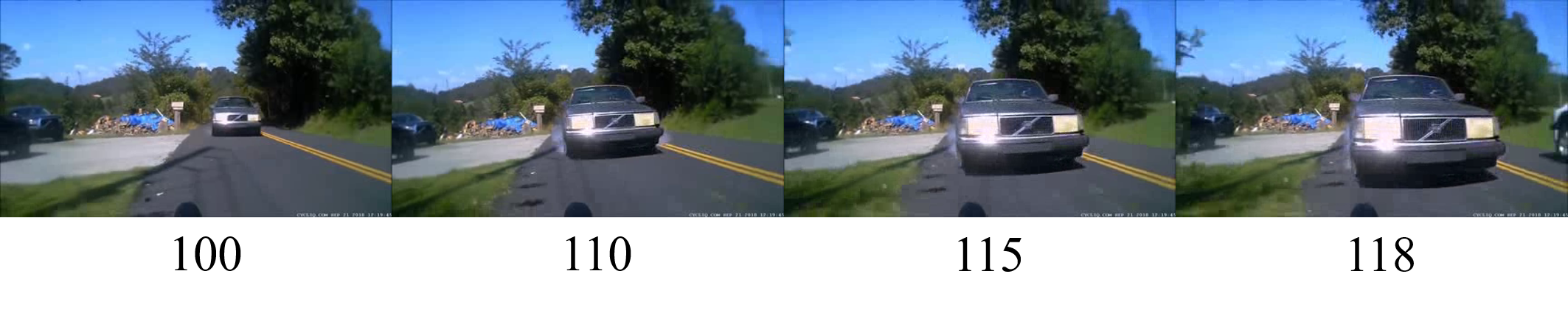}\vspace{1pt}
		\includegraphics[width=1\linewidth]{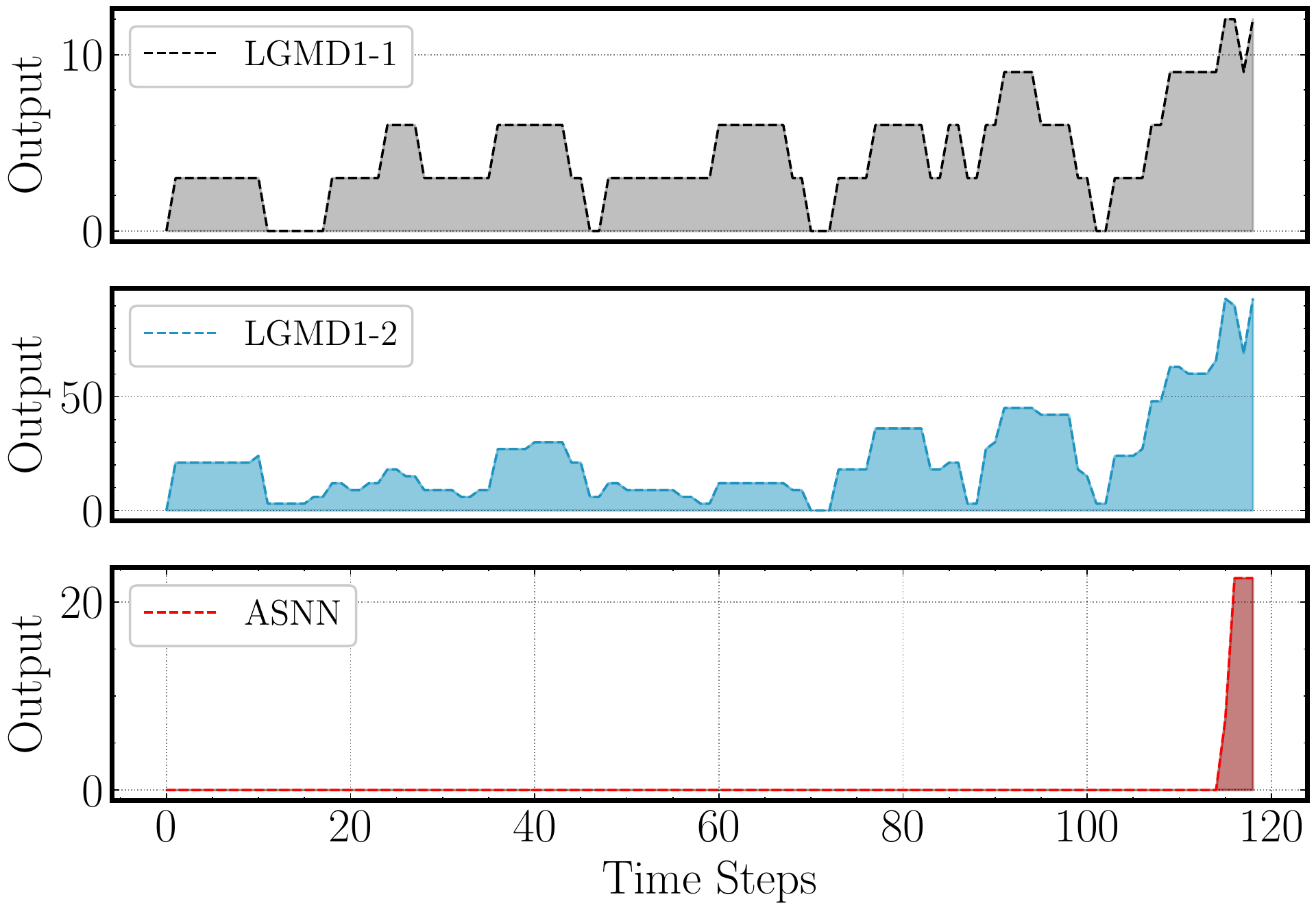}
		\end{minipage}}
	\subfigure[]{
    	\label{fig:s04f06i}
		\begin{minipage}[b]{0.3\linewidth}
		\includegraphics[width=1\linewidth]{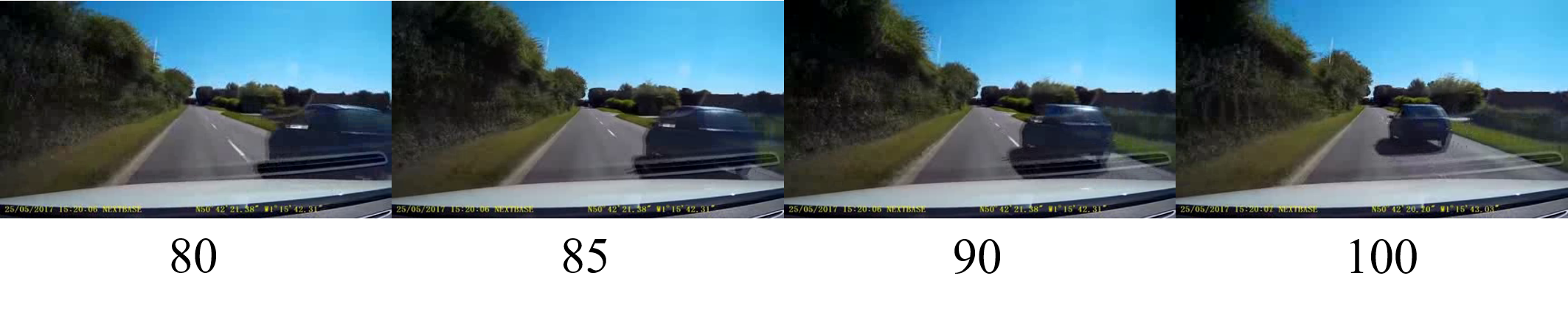}\vspace{1pt}
		\includegraphics[width=1\linewidth]{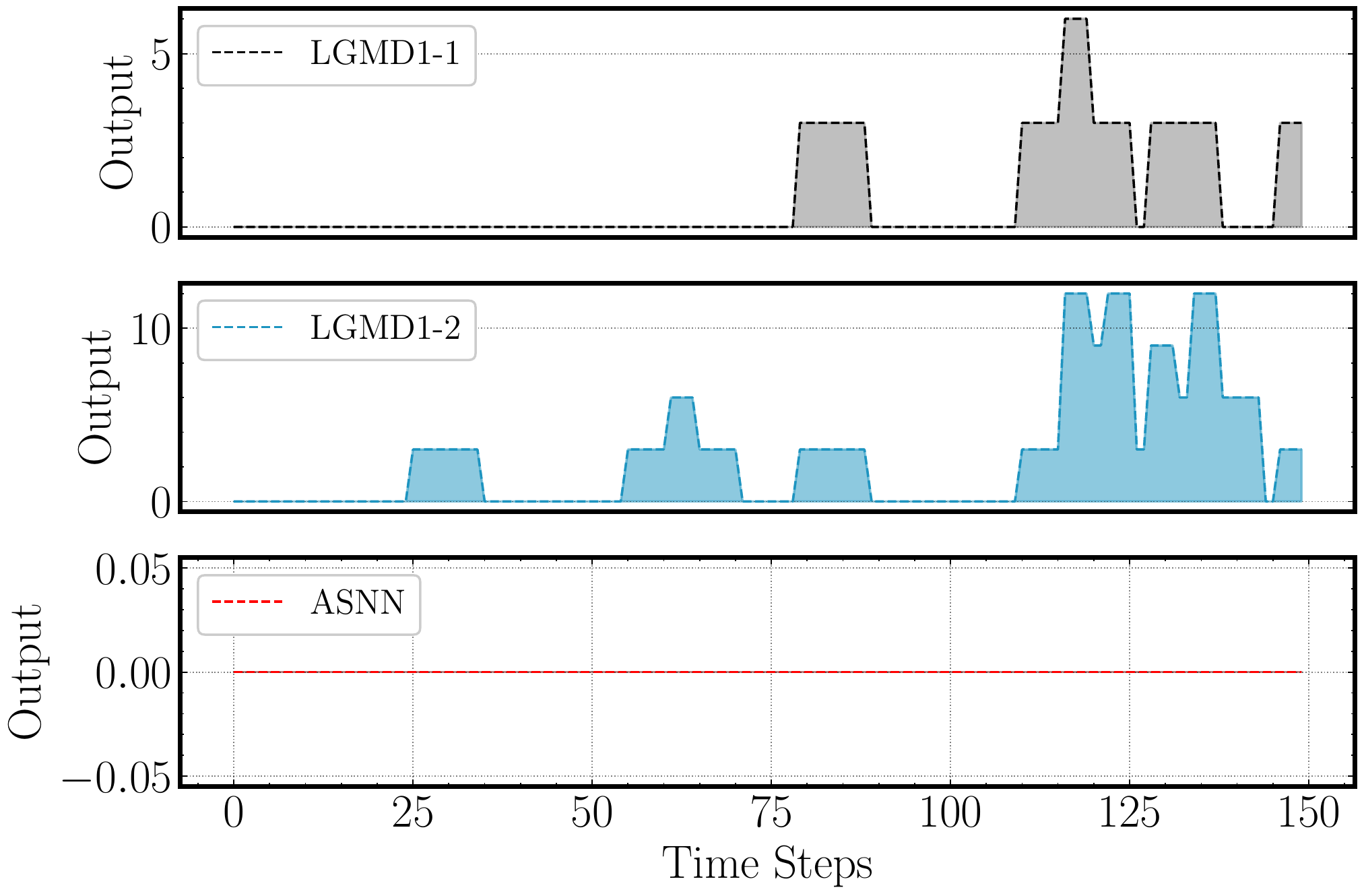}
		\end{minipage}}
    \caption{The neural responses of the proposed ASNN and the comparative LGMD model in three non-collision scenarios.}
    \label{fig:s04f07}
\end{figure*}

In this experiment, $K_{sp}$, $T_{sp}$, $N_t$ and $T_c$ are set to 10, 0.7, 4 and 1 respectively. Other parameters are same as the above synthetic experiment. The results of neural responses are depicted in Fig.\ref{fig:s04f06}. In this group of tests, all vehicles have collided. The proposed model works stably and effectively to detect potential collisions timely in various complex environments. The spike frequency increases dramatically only before truth colliding moments, and remains silent at other times. The LGMD1-1 and LGMD1-2 also increase the spike frequency before collision risks. However, the greatest weakness is not being able to keep silent in some non-collision situations. Especially even if some targets translate fastly, the responses of these two models will increase as well. Failure to effectively suppress background movement results in many false positives of collision detection. While, in this paper, the mechanism of lateral motion inhibition and directionally selctive inhibition can effectively filter out non-collision situations, which is also reflected in Fig.\ref{fig:s04f07}. For example, the camera shakes slightly in Fig.\ref{fig:s04f07}(b) such that the previous models are affected severely. While the proposed model is able to reduce the sensitivity to irrelevant background motion and only respond to the approaching targets.

\begin{figure*}[!htbp]
    \centering
    \includegraphics[width=0.9\textwidth]{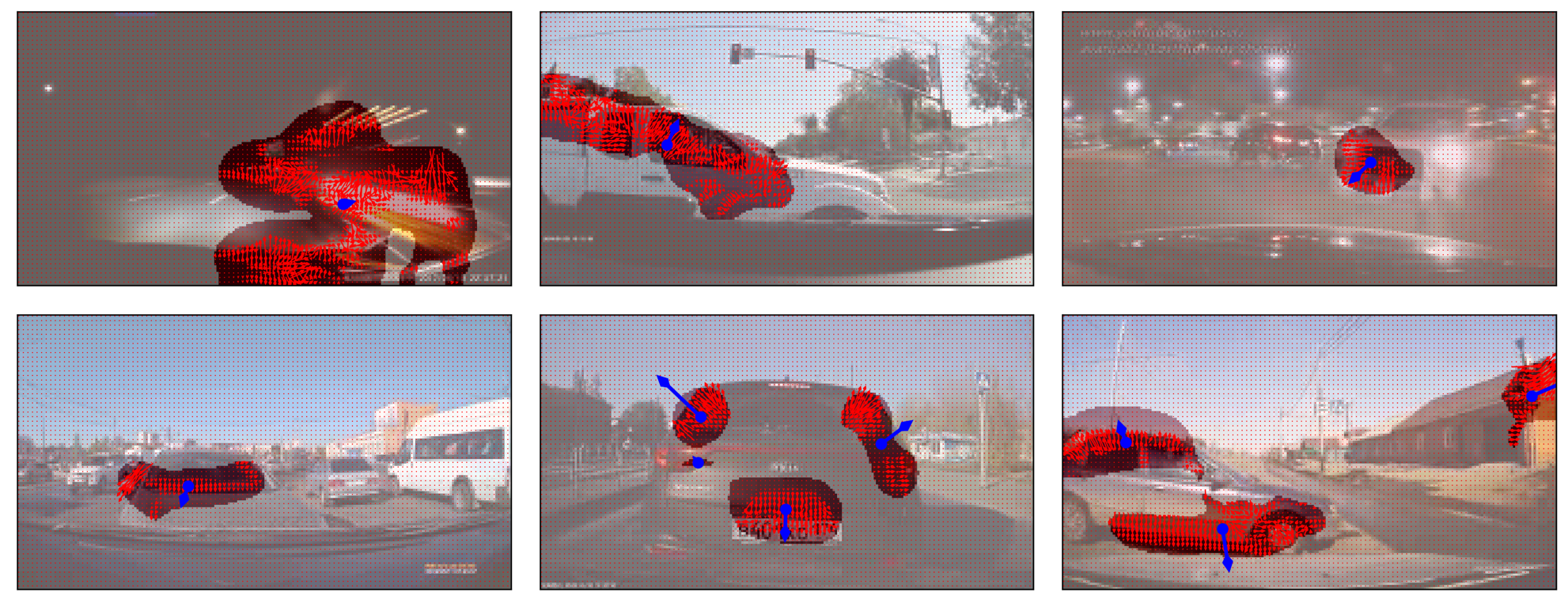}
    \caption{The estimated position and motion energy of the colliding targets in above six collision scenarios.}
    \label{fig:s04f08}
\end{figure*}

There is no doubt that extracting more motion information is crucial for the robot to make further decisions. In addition to assessing the risk of collision, our proposed model can also estimate the position and motion energy of the colliding targets. In above six collision scenarios, the looming regions are shown in Fig.\ref{fig:s04f08}. Obviously, all colliding cars are correctly detected before truth colliding moments. Meanwhile, the positions and directions of possible collision targets are computed, which can guide reflex control of the current mobile platform. The length of the blue arrow represents the motion energy of the looming region, which is obtained by a aforementioned population coded algorithm. It is closely related to the contrast between target and background, and the motion velocity of the looming target. Thus, compared with the previous models, our proposed model can not only detect the coming collision events accurately and robustly, but also extract the position and motion energy of the colliding targets.

\subsection{Robot Tests}
\begin{figure}[!htbp]
    \centering
    \includegraphics[width=0.3\textwidth]{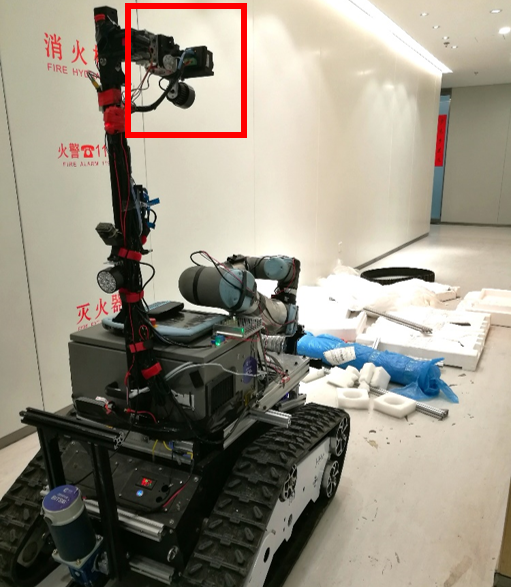}
    \caption{The scene of real robotic implementation.}
    \label{fig:s04f09}
\end{figure}

In this section, we further investigated the performance of collision detection for real robotic implementation. The backgrounds of most of the above on-road driving scenarios change smoothly and continuously when vehicles are driving. However, in some challenging natural environments such as rough terrains or in some highly dynamic tasks, the background of visual input is usually unstable and moving quickly due to the shaking camera or the rotating robot platform. For example, our platform is a tracked robot as shown in Fig.\ref{fig:s04f09}. Unlike wheeled robots with better motion stability, it usually shakes due to the frequent friction between the track and the ground. We compare the neural responses of our proposed model with the previous models in a general task, including turning, moving forward and collision detection. A low-cost camera is used to collect the real-time RGB image sequences, where the FPS is 25Hz. The size of each image is reshape as $320\times 240$ pixels. The weight coefficients are set as $[v_{+},v_{-},w_{+}^1,w_{-}^2,w_{-}^1,w_{+}^2]=[1.0,1.0,1.0,0.4,1.0,0.4]$. $K_{sp}$, $T_{sp}$, $N_t$ and $T_c$ are set to 20, 0.52, 4 and 1 respectively. Other parameters are the same as ones in the above vehicle tests.

\begin{figure*}[!htbp]
    \centering
    \includegraphics[width=0.8\textwidth]{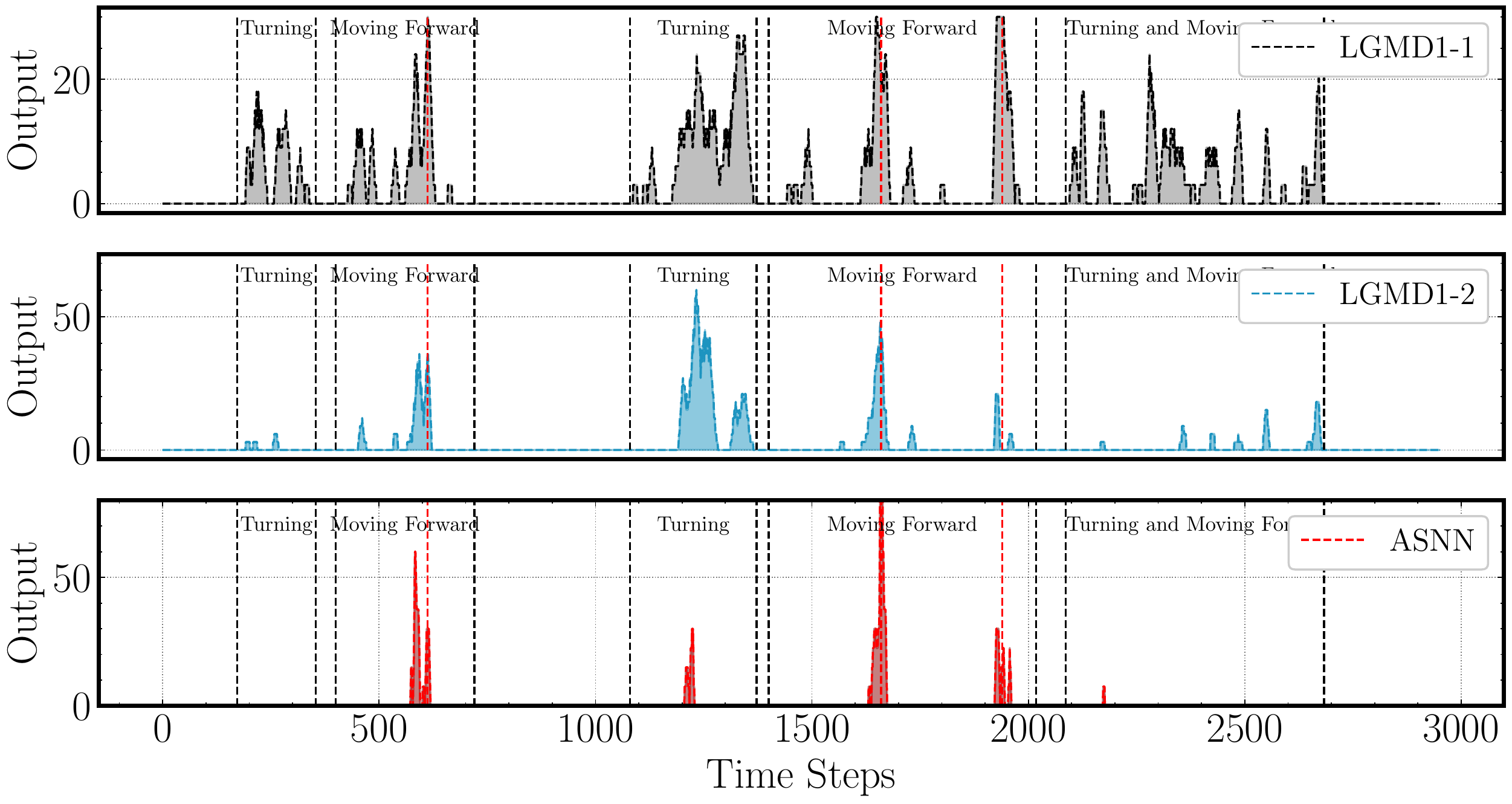}
    \caption{The collision detection of a tracked robot running over the flat terrain.}
    \label{fig:s04f10}
\end{figure*}

In the first case, the terrain is flat such that the camera shakes not too much. In this episode, we mainly investigate the influence of unstable movement and turning. The spike frequencies of three models are shown in Fig.\ref{fig:s04f10}. The result demonstrates that the proposed neural model is able to detect collision more accurately and robustly than the LGMD model, which is reflected in two aspects. Firstly, the proposed model can distinguish radial and lateral motion successfully such that it keeps silent for most translational motions of backgrounds. During turning, for example, the LGMD1-1 completely fails to filter out the translational features. The LGMD1-2 can suppress some noise effectively, but still can be affected by rapid turning, such as the response from 1080 to 1372 steps. One reason of it may be that spike frequency adaption in LGMD1-2 mediates the sensitivity to approach only based on sum of activations of all neurons instead of motion information of each pixel. It is easily affected by the contrast change of the background as well as the speed change. While the proposed model weakens the traslational motion through computing the motion information of the background, which enables it to surpress the background noise more robustly and reduce false positives. Secondly, the ASNN is able to detect the approaching objects accurately. The neurodynamics-based temporal filter and the push-pull computing structure formed by ON and OFF pathways contribute to improving the sensitivity of looming targets online from complex backgrounds, and increasing true positives. 

\begin{figure*}[!htbp]
    \centering
    \includegraphics[width=0.95\textwidth]{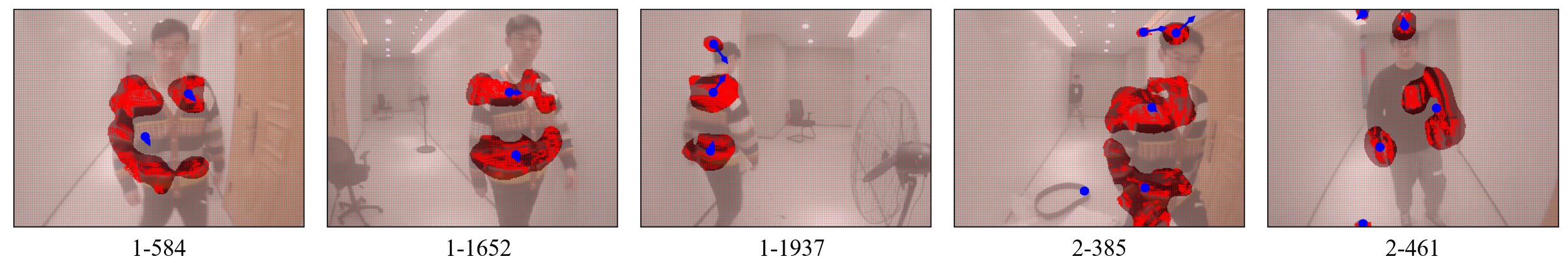}
    \caption{The estimated position and motion energy of the colliding targets on the flat-terrain (1-584, 1-1652 and 1-1937) and rough-terrain (2-385 and 2-461) situation.}
    \label{fig:s04f11}
\end{figure*}

Similar to the previous experiment, we also extract the motion information before potential colliding moments, as shown in Fig.\ref{fig:s04f11}. The position and motion direction of the colliding regions are estimated correctly, which is conducive to guiding the reflex control. For example, the target approaches from right side at step 584, which is able to trigger the robot to evade to the left. About the reflex control, we will study further in future work. At step 1937, the person actually moves away from the robot. A potential collision event is detected since the robot moves forward as well such that the relative distance between the two is decreasing. However, the magnitude of the response is relatively small, compared with the approach process. 

\begin{figure*}[!htbp]
    \centering
    \includegraphics[width=0.8\textwidth]{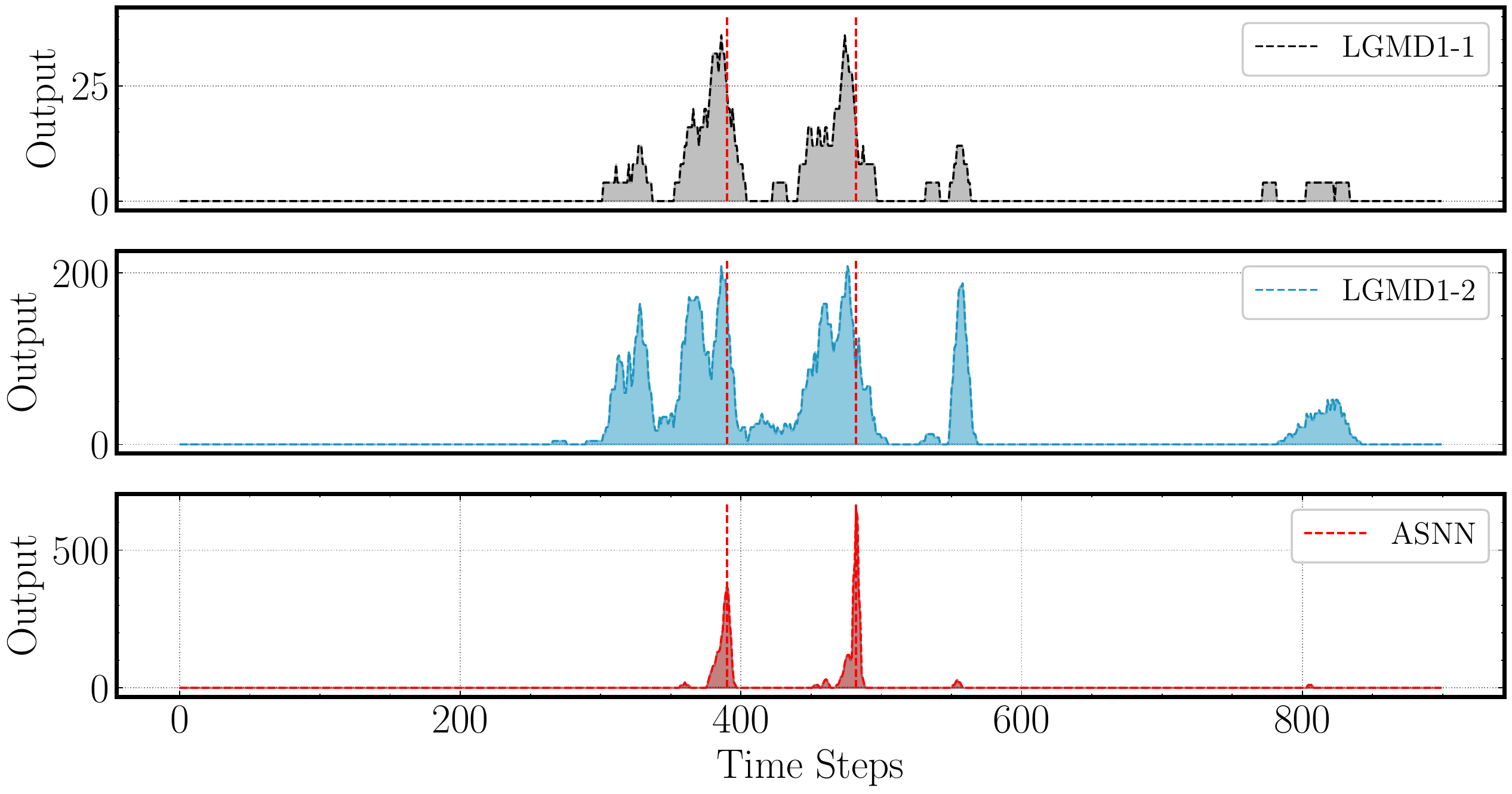}
    \caption{The collision detection of a tracked robot running over the rough terrain.}
    \label{fig:s04f12}
\end{figure*}

After that, we evaluate the performance of collision detection on rough terrain (Fig.\ref{fig:s04f12}), where the camera shakes strongly. Two people approach the robot one after another when the robot passes over this section of the road (Fig.\ref{fig:s04f11}). Fig.\ref{fig:s04f12} records the spike frequencies of three models. Obviously, the shaky background brings great challenge to LGMD1-1 and LGMD1-2. It is difficult for them to distinguish looming objects from a highly dynamic background. While the ASNN identifies the approaching targets accurately and suppresses irrelevant motion successfully. Herein, we choose the direction with the largest motion energy (top-1 direction) in the course of directionally selective attention. It is worth noting that more finely directionally selective spatial filtering will improve the performance of direction estimation and background suppression further, but it will increase the computational burden at the same time. 


\section{CONCLUSION}
Detecting approaching objects accurately and robustly is not only essential for animals to survive in nature, but also crutial and challenging for robotic vision systems to perform collision detection and obstable avoidance. Inspired by the neural circuit of elementary motion vision in the mammalian retina, this paper proposes a bioinspired approach-sensitive neural network model to improve the performance of collision detection. Three contributions are included. Firstly, a direction-selective visual processing module is built based on a spatiotemporal energy model, which is able to estimate motion direction accurately. Then, a novel approach-sensitive neural network is proposed. It is a push-pull structure formed by ON and OFF pathways, which responds strongly to approaching motion rather than lateral motion. Finally, we introduce a mechanism of attention, including approach-sensitive attention and directionally selective attention, which can suppress the translational motion of cluttered backgrounds effectively. In the experiment, we evaluate the approach sensitivy in synthetic scenarios, real driving scenarios and real robotic platform. The results show that the proposed model is able to detect collision accurately and robustly, especialy in cluttered and dynamic backgrounds. Meanwhile, more motion information like position and direction, is also extracted further for guiding rapid decision making. Thus, the proposed model provides an effective vision-based collision detection method for real robotic application, and has potential significance for neuroscience research.

\section{APPENDIX}
\subsection{Analysis of the neurodynamics-based temporal filtering} The response of this model (Eqn.5 and Eqn.6) to an instantaneous unit impulse can be calculated simply. For the dynamic, if $p$ is written for $\mathrm{d}/\mathrm{d}t$, the equation relating output $z_n$ to input $z_0$ is 
\begin{equation}\label{eqns3h5}
\left(p+\frac{A}{\tau}\right)^nz_n=\left( \frac{C}{\tau} \right) z_0.
\end{equation}
When $z_0$ is unit impulse signal, the solution of above equation is
\begin{equation}\label{eqns3h6}
z_n=\left( \frac{C}{\tau} \right)^n\frac{t^{n-1}e^{\frac{-At}{\tau}}}{(n-1)!}.
\end{equation}
Since output of temporal filtering $\mathcal{L}$ is the difference between $n$th and $(n+m)$th layers of neurons, the unit impulse response of .. is
\begin{equation}
\begin{split}
\mathcal{L}&=K\left( \frac{C}{\tau} \right)^n\frac{t^{n-1}e^{\frac{-At}{\tau}}}{(n-1)!}-K\left( \frac{C}{\tau} \right)^{n+m}\frac{t^{n+m-1}e^{\frac{-At}{\tau}}}{(n+m-1)!}\\
&=K\left(\frac{C}{\tau}\right)^nt^{n-1}e^{\frac{-At}{\tau}}\left[ \frac{1}{(n-1)!}-\frac{\left( \frac{Ct}{\tau}\right)^m}{(n+m-1)!}\right].
\end{split}
\end{equation}
Let $a=\frac{A}{\tau}$, $b=\frac{C}{\tau}$, then we have 
\begin{equation}
\mathcal{L}=Ke^{-at}\left[ \frac{b^nt^{n-1}}{(n-1)!} - \frac{b^{n+m}t^{n+m-1}}{(n+m-1)!} \right].
\end{equation}
The derivative with respect to $t$ is 
\begin{equation}
\begin{split}
\frac{\mathrm{d}\mathcal{L}}{\mathrm{d}t}&=-Kae^{-at}\left[ \frac{b^nt^{n-1}}{(n-1)!} - \frac{b^{n+m}t^{n+m-1}}{(n+m-1)!} \right]\\
&\quad +Ke^{-at}\left[ \frac{b^nt^{n-2}}{(n-2)!}-\frac{b^{n+m}t^{n+m-2}}{(n+m-2)!} \right].
\end{split}
\end{equation}
Let $\frac{\mathrm{d}\mathcal{L}}{\mathrm{d}t}=0$, we have 
\begin{equation}
\begin{split}
\left[ \frac{b^nt^{n-2}}{(n-2)!}-\frac{b^{n+m}t^{n+m-2}}{(n+m-2)!} \right]
-a\left[ \frac{b^nt^{n-1}}{(n-1)!} - \frac{b^{n+m}t^{n+m-1}}{(n+m-1)!} \right]=0.
\end{split}
\end{equation}
Simplify this equation, there is 
\begin{equation}
\frac{n-1-at}{(n-1)!}-\frac{b^mt^m(n+m-1-at)}{(n+m-1)!}=0.
\end{equation}
Then we have
\begin{equation}
\begin{split}
&ab^m(n-1)!t^{m+1}-b^m(n+m-1)(n-1)!t^m\\
&-a(n+m-1)!t+(n-1)(n+m-1)!=0.
\end{split}
\end{equation}
Especially when $m=1$, we have
\begin{equation}
ab(n-1)!t^2-(a+b)n!t+(n-1)n!=0.
\end{equation}
Since 
\begin{equation}
\begin{split}
\Delta &=\left[ (a+b)n! \right]^2 -4ab(n-1)! \cdot (n-1)n!\\
&=(an!)^2+(bn!)^2+2ab(n!)^2-4ab(n-1)!\cdot (n-1) n! \\
&>(an!)^2+(bn!)^2+2ab(n!)^2-4ab(n!)^2\\
&=\left[ (a-b)n! \right]^2 \geq 0,
\end{split}
\end{equation}
there are two real roots.
\begin{equation}
t_1=\frac{(a+b)n!-\sqrt{\Delta}}{2ab(n-1)!},
\end{equation}
\begin{equation}
t_2=\frac{(a+b)n!+\sqrt{\Delta}}{2ab(n-1)!}.
\end{equation}
The activity reaches a maximum at time $t_1$, and reaches a minimum at time $t_2$. When $A>C$ ($a>b$), then $t_1<\frac{n}{a}$ and $t_2>\frac{n}{b}$. Otherwise when $A<C$ ($a<b$), then $t_1<\frac{n}{b}$ and $t_2>\frac{n}{a}$.


%

%


\ifCLASSOPTIONcaptionsoff
  \newpage
\fi



\bibliographystyle{IEEEtran}
\bibliography{ref}
%
%





\end{document}